\documentclass[final]{elsarticle}
\usepackage{framed,multirow}
\usepackage{geometry}
\makeatletter
\def\ps@pprintTitle{%
	\let\@oddhead\@empty
	\let\@evenhead\@empty
	\let\@oddfoot\@empty
	\let\@evenfoot\@oddfoot
}
\makeatother

\usepackage{float}
\usepackage{hyperref}
\usepackage{amsmath}
\usepackage{amssymb}
\usepackage{bbm}
\usepackage{nameref}
\usepackage{etoolbox}
\usepackage{url}            
\usepackage{booktabs}       
\usepackage{amsfonts}       
\usepackage{nicefrac}       
\usepackage{microtype}      
\usepackage{rotating}
\usepackage{multirow}
\usepackage{graphicx} 
\usepackage{graphicx}  
\newtheorem{thr}{Theorem}
\newtheorem{defi}{Definition}
\newtheorem{pro}{Proposition}
\usepackage{todonotes}
\usepackage[linesnumbered,ruled,vlined]{algorithm2e}
\SetCommentSty{mycommfont}

\SetKwInput{KwInput}{Input}                
\SetKwInput{KwOutput}{Output}              

\usepackage{accents}

\begin{document}

\begin{frontmatter}

\title{TATL: Task Agnostic Transfer Learning for Skin Attributes Detection}

\author[1,2]{Duy M. H. Nguyen}
\author[3]{Thu T. Nguyen}
\author[4]{Huong Vu}
\author[5]{Quang Pham}
\author[6]{Manh-Duy Nguyen}
\author[7,8,9]{Binh T. Nguyen \corref{mycorrespondingauthor}}
\author[1,10]{Daniel Sonntag}

\cortext[mycorrespondingauthor]{Corresponding author: ngtbinh@hcmus.edu.vn (Binh T. Nguyen)}

\address[1]{German Research Center for Artificial Intelligence, Saarbrücken, Germany}
\address[2]{Max Planck Institute for Informatics, Germany}
\address[3]{University of Louisiana at Lafayette, USA}
\address[4]{University of California, Berkeley, USA}
\address[5]{School of Computing and Information Systems, Singapore Management University}
\address[6]{School of Computing, Dublin City University, Ireland}
\address[7]{AISIA Research Lab, Ho Chi Minh City, Vietnam}
\address[8]{University of Science, Ho Chi Minh City, Vietnam}
\address[9]{Vietnam National University, Ho Chi Minh City, Vietnam}
\address[10]{Oldenburg University, Germany}

\date{}
\begin{abstract}
Existing skin attributes detection methods usually initialize with a pre-trained Imagenet network and then fine-tune on a medical target task. However, we argue that such approaches are suboptimal because medical datasets are largely different from ImageNet and often contain limited training samples. In this work, we propose \emph{Task Agnostic Transfer Learning (TATL)}, a novel framework motivated by dermatologists' behaviors in the skincare context. TATL learns an attribute-agnostic segmenter that detects lesion skin regions and then transfers this knowledge to a set of attribute-specific classifiers to detect each particular attribute. Since TATL's attribute-agnostic segmenter only detects skin attribute regions, it enjoys ample data from all attributes, allows transferring knowledge among features, and compensates for the lack of training data from rare attributes. 
We conduct extensive experiments to evaluate the proposed TATL transfer learning mechanism with various neural network architectures on two popular skin attributes detection benchmarks.
The empirical results show that TATL not only works well with multiple architectures but also can achieve state-of-the-art performances, while enjoying minimal model and computational complexities. We also provide theoretical insights and explanations for why our transfer learning framework performs well in practice.
\end{abstract}


\end{frontmatter}

\section{Introduction}\label{sec:intro}
Melanoma is one of the most dangerous types of skin cancer. Even though it only accounts for 1\% of all skin cancer cases, it is responsible for the majority of skin cancer deaths \citep{ward2017cutaneous}. In 2021, it is estimated that there will be 207,390 new cases of melanoma will be diagnosed and 7,180 recent deaths from the disease in the United States alone \citep{Cancer2021}.

Moreover, the 5-year relative survival rate for melanoma reduced from 99\% for cases diagnosed at a localized stage to 27\% for a distant stage \citep{Cancer2021}. Therefore, there have been tremendous efforts in detecting the disease in its early stages \citep{masood_computer_2013,curiel-lewandrowski_artificial_2019}. One of the most promising technology is \emph{Dermoscopy}, which can generate high-resolution images of skin lesions  and allows dermatologists to examine the lesion regions more carefully \citep{celebi_dermoscopy_2019}. However, dermoscopy still requires extensive training, which is expensive, time-consuming, error-prone, and might not be widely available \citep{zalaudek2008time}. Therefore, it is important and highly beneficial to develop automatic systems to detect abnormal skin lesions and aid dermatologists during diagnosis \citep{nunnari2021overlap}. 

\begin{figure}[!hbtp]
\begin{center}
	\includegraphics[width=0.25\textwidth]{images/ISIC_0000013.jpg}
	\includegraphics[width=0.25\textwidth]{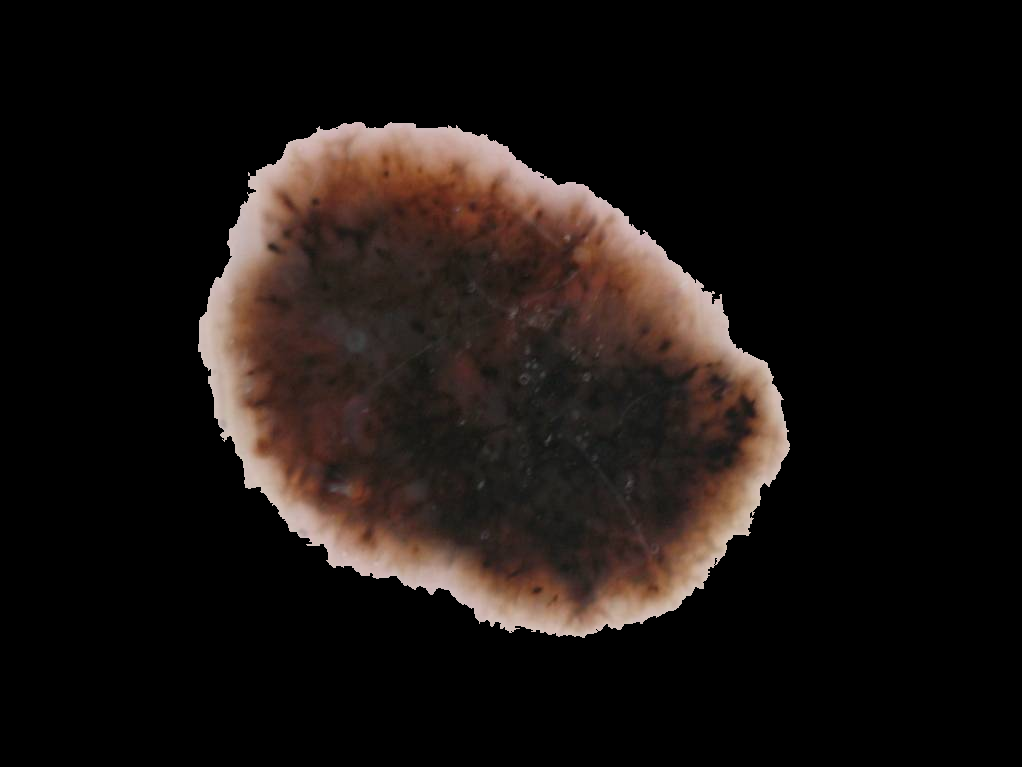}
	\includegraphics[width=0.25\textwidth]{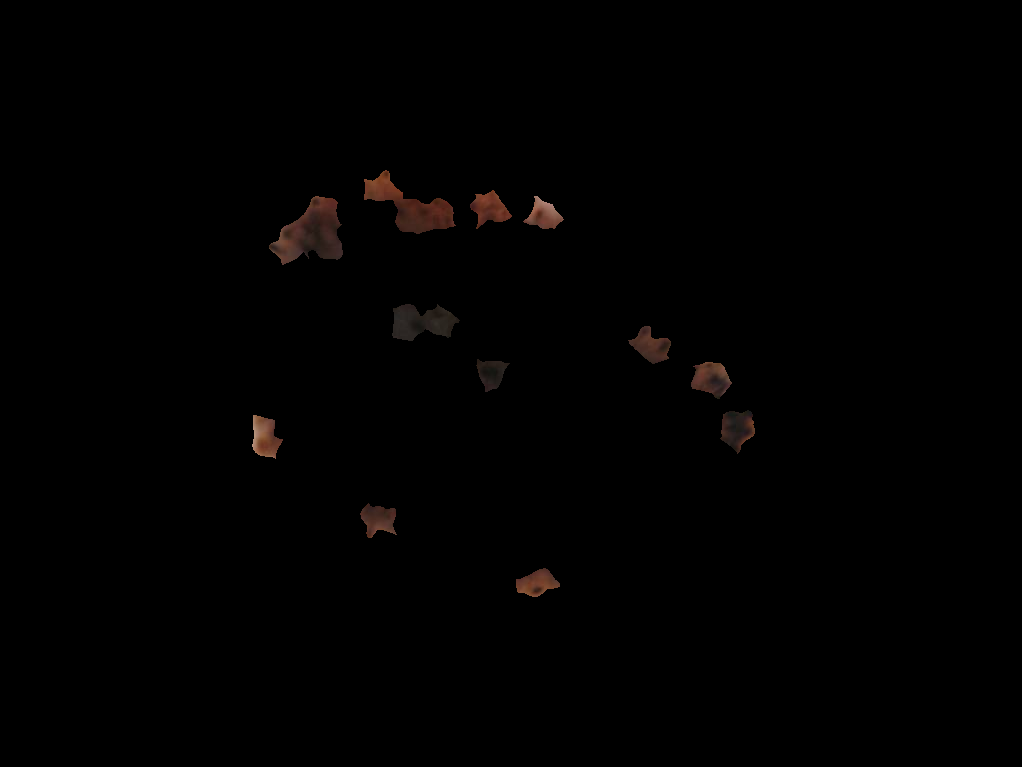}\\
	\includegraphics[width=0.25\textwidth]{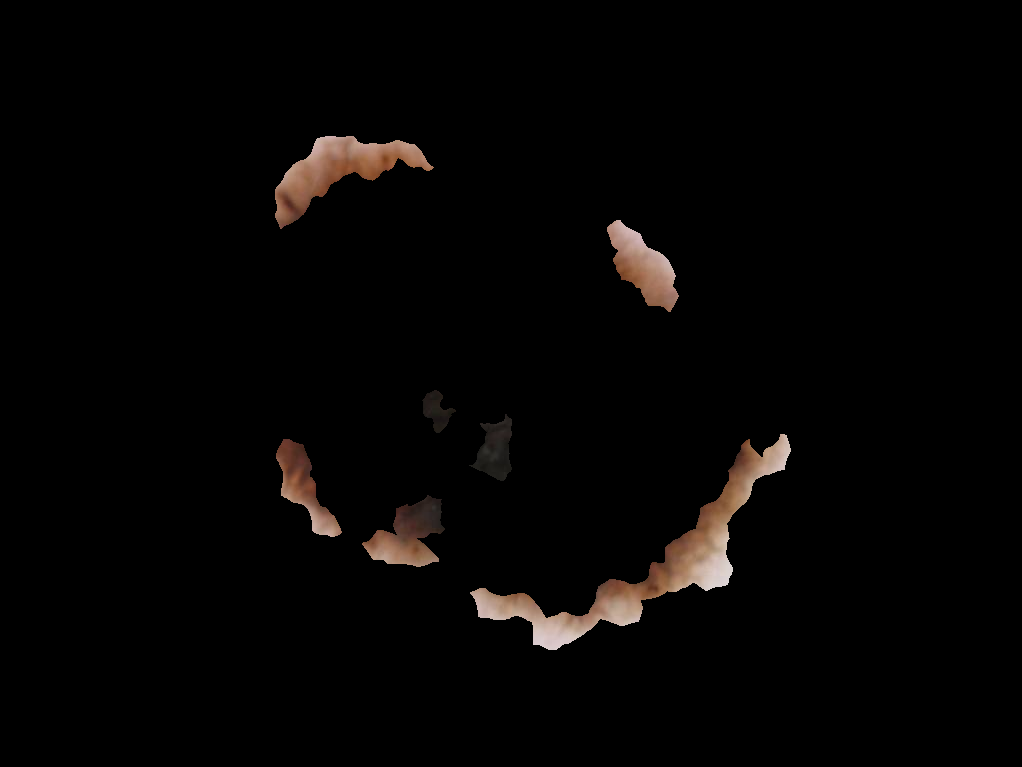}
	\includegraphics[width=0.25\textwidth]{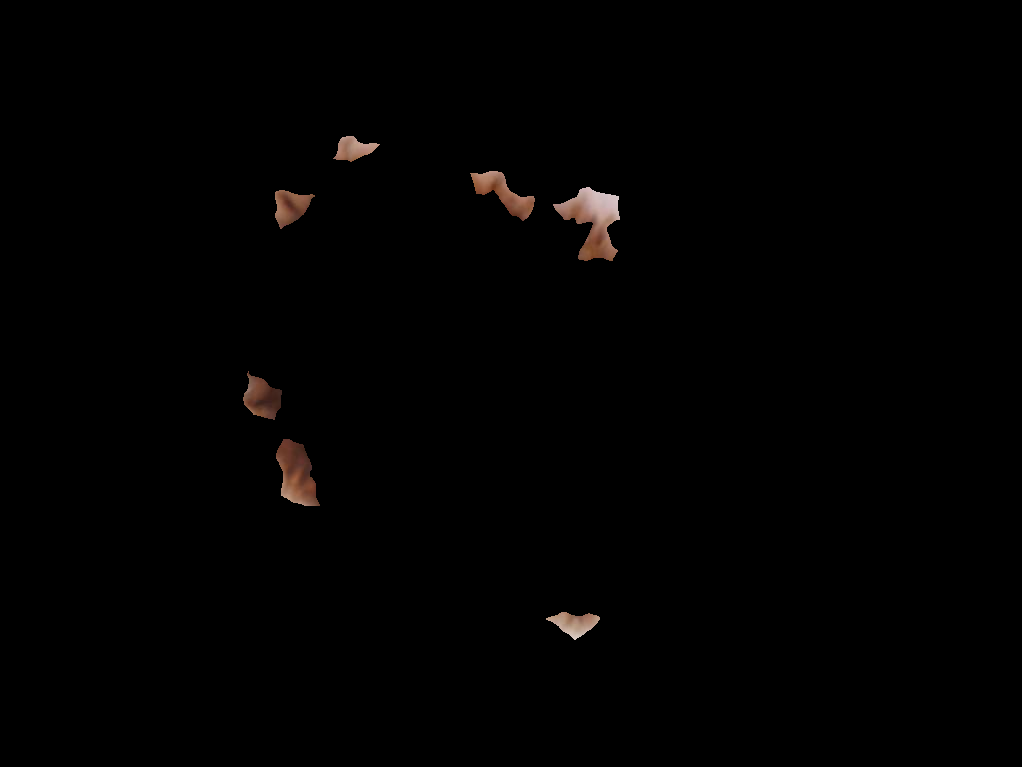}
	\includegraphics[width=0.25\textwidth]{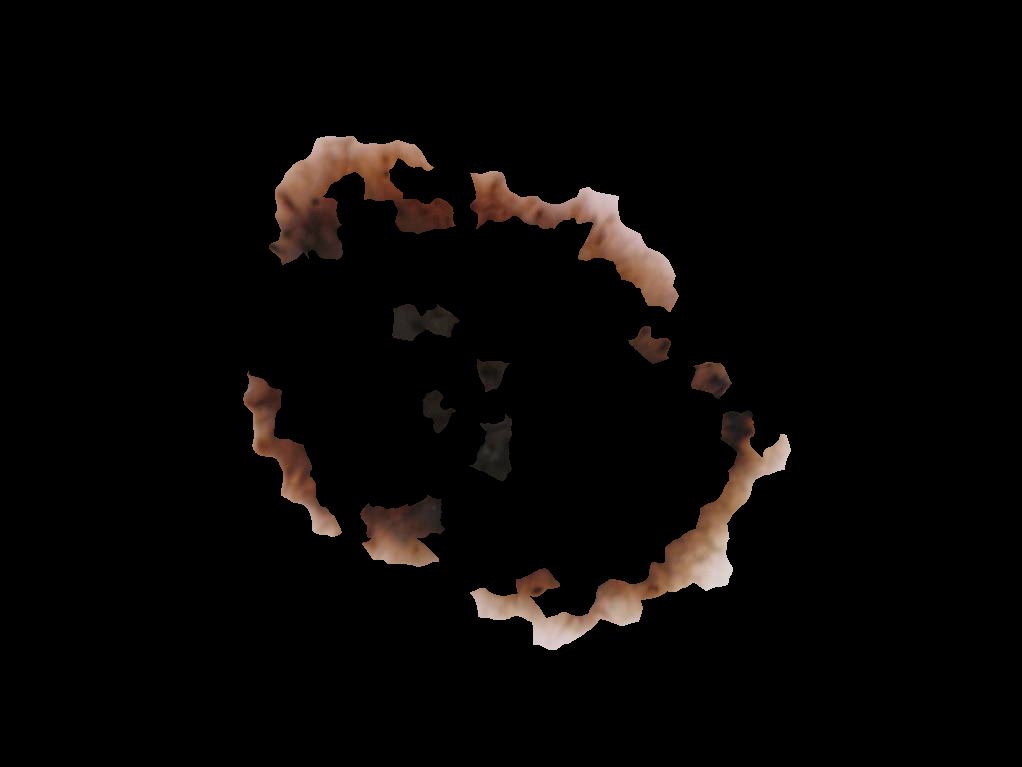}
	\caption{From left to right: the picture of melanoma, its segmentation, and the annotation for globules (upper row), the annotation for Pigment Network, Streaks, and the three's union (below row). The images are taken from ISIC 2018 \citep{codella2019skin}.}
	\label{fig:lesion-examples}
\end{center}
\end{figure}

For this purpose, the International Skin Imaging Collaboration (ISIC) hosted challenges for automatic melanoma detection based on dermoscopic images (ISIC-2018, ISIC-2017)~\citep{codella2019skin,codella2018skin}. 
In this work, we focus on Task 2 of predicting the locations of dermoscopic attributes in an image. In particular, there are \emph{five} dermoscopic attributes that the challenge focused on: Streaks, Globules, Pigment Network, Negative Network, and Milia-like Cysts. Locating these clinically meaningful skin lesion patterns helps detect anomalous regions and provides an explanation for dermatologists to verify and make further diagnoses. For instance, the Negative Network, which consists of relatively light regions and some darker regions, is usually considered a melanoma-specific structure \citep{pizzichetta2013negative}. We provide an example of such attributes in Figure~\ref{fig:lesion-examples}. 

Due to its highly competitive nature, many methods in the ISIC challenges are based on the ensemble strategies with different types of ImageNet (ImgNet) pre-trained backbones. For instance, the ISIC 2018's winner method \citep{koohbanani2018leveraging} employed a mixture of four pre-trained networks ResNet152~\citep{he2016deep}, DenseNet169~\citep{huang2017densely}, Xception~\citep{chollet2017xception}, and ResNetV2~\citep{szegedy2017inception} in the encoding part of the U-Net~\citep{navab_u-net:_2015} segmentation models and then performed transfer learning on the ISIC 2018 training set.  
Although \citet{koohbanani2018leveraging} achieved state-of-the-art performance, it is not an attractive method in practice because of \emph{two} reasons. 
First, \citet{koohbanani2018leveraging} simultaneously stores five different models in a single GPU in the pre-training phase, which costs a total of over 300 million parameters (Table~\ref{tab:parameter}). As a result, this approach cannot be deployed on most {\bf non} high-end GPU cards, which limits its accessibility to practical deployment and its extensibility for future research.
Second, recent works in~\citet{raghu2019transfusion, cheplygina2019cats, nguyen2020visually} showed that transfer learning from ImageNet might be sub-optimal in many scenarios because medical images are primarily different from ImageNet data. Moreover, most medical datasets, including ISIC-2018 and ISIC-2017, suffer from the scarcity of training data, and the number of instances per attribute is imbalanced, as depicted in Table \ref{tab:descr}.

To address the problems of massive memory requirement and domain gap between ImageNet and medical images mentioned above, we propose \emph{Task Agnostic Transfer Learning} (TATL), an efficient framework to detect skin attributes in dermoscopic images. TATL's design is inspired by how dermatologists diagnose in practice: identify abnormal skin regions in the first step and then inspect them more closely in the second step \footnote{\url{https://www.polyclinic.com/health-wellness-library/find-skin-cancer.html}}$^{,}$\footnote{\url{https://www.cancer.org/content/dam/CRC/PDF/Public/8825.00.pdf}}.
Unlike previous works that try to segment skin attributes on the image directly, TATL introduces an \textit{Attribute-Agnostic Segmenter} that first detects anomalous regions in an image, regardless of their attributes. Then, TATL transfers this segmenter's knowledge to a set of attributes-specific segmenters \textit{(Target Segmenters)} to detect each specific attribute. Notably, the Attribute-Agnostic Segmenter is task-agnostic because it only identifies abnormal skin regions, possibly including data from all attributes. Therefore, TATL alleviates the lack of training samples by training the segmenter in the first stage. Furthermore, through the experiments, we find out that TATL also facilitates knowledge sharing among the attribute-segmenters, thus enhancing the generalization and stability of the whole system.
Furthermore, we also provide theoretical insights reasoning that TATL works by bridging the gap between the target task's data and the source dataset. The attribute-specific classifiers are particularly initialized from the TATL's Union-Segmenter, which enjoys a tighter generalization error bound than other methods initializing from ImageNet. This analysis sheds light on the promising performances of TATL.

In summary,  our contributions are threefold. Firstly, we propose TATL, a novel transfer learning approach for skin attribute detection. Extensive experiments on the ISIC 2018 and ISIC 2017 benchmarks validate the effectiveness of TATL against state-of-the-art methods while requiring only 1/30 number of parameters compared with the ISIC 2018's winner. Furthermore, TATL significantly improves several skin diagnosis methods pre-trained on ImageNet, especially for attributes with limited training instances. Secondly, we provide theoretical insights explaining the success of TATL; thereby, TATL has a high ability to reduce domain gaps by shifting from color images (ImageNet) to the medical domain (Skin images). Finally, TATL can provide informative outputs to aid dermatologists in further diagnosing as it can simultaneously predict both abnormal regions and different kinds of skin features, which makes a promising first step towards a more practical and beneficial medical-aid deep learning system. 
\section{Related Work}
\subsection{Transfer Learning for Medical Image Analysis}
Medical image analysis is a vital research venue and has a significant impact on practice. However, most medical image datasets {have limited} training {samples} and often suffer from {the} data-imbalanced {problem}. Therefore, a popular strategy is \emph{transfer learning}, which uses a pre-trained ImageNet model as an initialization to build additional components. Transfer learning is a base of many existing methods~\citep{abramoff2016improved, de2018clinically, gulshan2016development,rajpurkar2017chexnet}, and is a norm for practitioners \citep{tan2018survey}.
However, recent studies in~\citet{cheplygina2019cats,he2020sample} conducted a large-scale analysis on the benefit of this strategy. They concluded that {for medical images,} transfer learning based on pre-trained ImageNet is not consistently better than random initialization. One reason is that medical images are vastly different from the ImageNet dataset, resulting in the pre-trained {weights that} are not helpful for the current task. Another reason is that medical data are often imbalanced and rare due to data privacy. For example, Table~\ref{tab:descr} illustrates the distributions of each skin characteristic in the ISIC 2017 and ICSIC 2018 datasets, with the rarest class (Streaks) accounting for only 7.98\% (113 images) and 3.86\% (100 images) of the training data, respectively. In comparison, the most common class (Pigment Network) accounts for 79.03\% (1119 images) and 58.67\% (1522 images) of the total samples. Fortunately, TATL can address the scarcity of training data in such situations by transferring the knowledge from the \textit{Attribute-Agnostic Segmenter}. Moreover, we since apply the strategy \textit{one class one model} \citep{buda2018systematic}, each \textit{Target-Segmenter} classifier in TATL only detects one attribute, and can alleviate the data's imbalance problem.
\subsection{Self-Supervised Learning In Computer Vision}
Self-supervised learning, first mentioned in \citet{schmidhuber1990making}, refers to a technique of creating additional tasks for training where the label is also a part of the data (images) rather than a set of separate labels (annotations). This strategy has been a successful pre-training technique in various vision applications, including image colorization \citep{vondrick2018tracking,larsson2016learning,zhang2016colorful}, image {impainting} \citep{pathak2016context, chen2020simple}, and video representation \citep{misra2016shuffle,lee2017unsupervised}. In self-supervised learning, a newly created task for pre-training is called the \textit{``pretext task"}, and the main tasks used for fine-tuning are called the \textit{``downstream tasks"}. Various strategies have been proposed to construct the pretext task based on the {image rotation \citep{DBLP:journals/corr/abs-1803-07728}}, temporal correspondence \citep{li2019joint,wang2019learning}, cross-modal consistency \citep{wang2019reinforced} and instance discrimination with contrastive learning \citep{wu2018unsupervised}. Recently, in the medical domain, \citet{he2020sample}  successfully applied self-supervised learning in diagnosing COVID19 from CT scans based on contrastive self-supervised learning for reducing the risk of overfitting. \citet{chen2019self} also presented a context restoration framework in which the image is disordered by randomly changing the order of their sub-patches, and then a neural network is trained to recover the original input. 

Our TATL approach shares some similarities with self-supervised learning in the sense of building better pretrained models without extra training instances or learning through auxiliary tasks. Here, the Attribute-Agnostic Segmenter plays a role as of learning a pretext task, while detecting the attributes in the Target Segmenters are learning the downstream tasks. {However, our work differs from the SSL approaches because we define an auxiliary task through solving $(x, g(y))$ while SSL methods follow the scheme $(x, f(x))$ where $(x, y)$ indicates for the image and corresponding label, $g$ is an operator on the training label, and $f$ is another transformation on the image such as rotation \citep{DBLP:journals/corr/abs-1803-07728}, dividing images into sub-patches and suffering their positions \citep{chen2019self}}.
The construction of $g$ is specifically designed for the medical domain, which makes TATL's pretext task closely complements the subsequent downstream tasks.  
Therefore, if the pretext task of recognizing {skin attribute} regions can perform well, it will likely facilitate detecting such areas' attributes. Finally, by providing skin attribute regions or abnormal regions from the pretext task, TATL is helpful to end-users by allowing dermatologists to validate the employed system's diagnostics.

\newcommand*\BitOr{\mathbin{|}}
\section{Preliminaries}
This section aims to formalize our problem setting and outline the dermatologists' practice to diagnose skin attributes, which later motivates our method. 

\subsection{Problem Statement and Background}
We consider the skin attributes detection problem on a target dataset consisting of training images and their corresponding masks {$\mathcal{D}$} $= \{(x_1, \{y_1^{(i)}\}_{i=1}^{|Y|} ) \ldots, (x_n, \{y_n^{(i)}\}_{i=1}^{|Y|})\}$. 
The detector, parameterized by W, can be initialized from a pre-trained model on another dataset, which we call the source dataset {$\mathcal{D}^{src}$}. Moreover, each training sample in the target domain consists of an image $x \in \mathbb{R}^{c\times w \times h}$ and a set of labels $\{\{y^{(i)}\}_{i=1}^{|Y|}, y^{(i)} \in \mathbb{R}^{w \times h}\}$, where $y^{(i)}$ is a binary mask indicating the skin region associated with the $i$-th attribute. In this work, we consider {\bf five} different {attributes}: Globules, Milia, Negative Network, Pigment Network, and Streaks, shorthanded as  $Y = \{G, M, N, P, S\}$, i.e., $|Y| = 5$. It is worth noting that each sample may not have all the attributes and the label for those missing attributes is the empty mask. The training process can be performed by minimizing the empirical risk: 
\begin{align}
    & \widehat{f}(\{x_j, \{y_j^{(i)}\} \}_{j=1}^n |W) =  \arg\min_f \, \frac{1}{n}\sum_{j=1}^{n} \sum_{i=1}^{|Y|} \lambda_1 {L_{Tversky}}(\widehat{y}^{(i)}_j, y^{(i)}_j) + \lambda_2 {L_{Jaccard}}(\widehat{y}^{(i)}_j, y^{(i)}_j),
    \label{eq:1}
\end{align}
where $\widehat{y}^{(i)}_j$ denotes the binary mask prediction of the network on a sample about the $i$-th attribute. For each attribute, {we use the Tversky loss $L_{Tversky}$, which is a generalization of Dice loss \citep{eelbode2020optimization,jadon2020survey,salehi2017tversky} and the soft Jaccard loss functions $L_{Jaccard}$ \citep{eelbode2020optimization,kawahara2018fully}}, to penalise the deviation between network's predictions and the ground-truths. Formally, these loss functions can be calculated as:
\begin{align}
\small
{
L_{Tversky}(\widehat{y}, y)} &= 1 - \frac{\alpha + \langle y,\hat{y}\rangle}{\alpha + \langle y,\hat{y}\rangle + \beta \langle 1-\hat{y},y\rangle + (1-\beta)\langle \hat{y},1-y \rangle},  \label{eqD} \\
{L_{Jaccard}(\hat{y},y)} &= 1 - \frac{\alpha + \langle y, \hat{y} \rangle}{\alpha + ||y||_{1} + ||\hat{y}||_{1} - \langle y, \hat{y} \rangle}. \label{eq:J}
\end{align}
Here, the prediction $\hat{y}$ and the ground-truth $y$ are first re-shaped into a vector form; $\langle\,, \,\rangle$ and $||.||_{1}$ are the inner product and $L^{1}$ norm respectively. {The parameter $\alpha$ is used to ensure that the loss functions are not undefined when division by zero or in case $y = \hat{y} = 0$. The parameter $\beta$ in other way controls the magnitude of false positives and false negatives. In our experiment,  we choose $\lambda_{1} = \lambda_2 = 0.5$ to balance the importance of two loss functions and parameters $\alpha = 1,\, \beta = 0.6$ through validation experiments}.

\subsection{Inspirations from Dermatologists' Behaviours}
\begin{figure*}[]
    \begin{center}
	\includegraphics[width=1.0\textwidth]{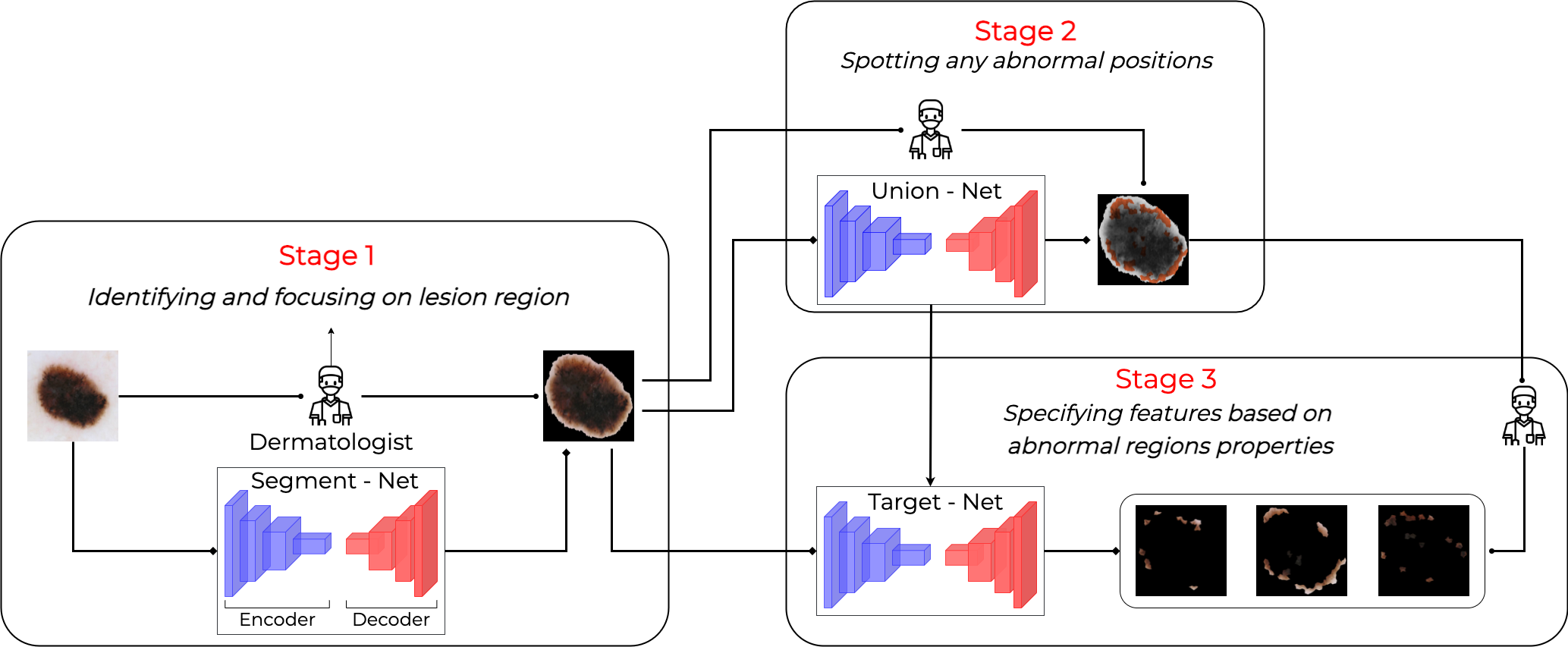}
	\caption{The prediction pipeline for each skin feature is motivated by a dermatologist's behavior when examining a patient.}
    \label{fig:prediction-method}
    \end{center}	
\end{figure*}
\label{sec:inspiration}

Figure \ref{fig:prediction-method} depicts a prediction pipeline inspired by the conventional diagnosis process of dermatologists, as discussed in Section \ref{sec:intro}. In the first step, dermatologists will identify lesion regions by eliminating irrelevant background and rescaling these regions to a higher resolution for better visualization (Stage 1). Following that, they continue to spot any abnormal and clinically relevant sub-areas on the lesion (Stage 2). Finally, by accounting for these factors, doctors diagnose specific skin attributes by comparing various features based on their textures and colors compared to nearby spaces (Stage 3). We argue that identifying lesion and skin attribute regions is crucial since it serves focal points for later steps, and we develop a skin attributes detection framework that closely follows the three-step procedure represented in Figure \ref{fig:prediction-method}.

We realize the diagnosis procedure into a single framework named  Task Agnostic Transfer Learning (TATL). First, {TATL} employs a Segment-Net to segment the lesion regions from an input image. Then, TATL trains an Attribute-Agnostic Segmenter to detect all {skin attribute regions} in the image, regardless of their attributes, which is inspired by the second step in the procedure. Finally, the parameters of the Attribute-Agnostic Segmenter are utilized as an initialization for the Target Segmenters (Tar-S), which are trained to identify just one specific attribute and are employed as the final step in the method.

TATL not only closely resembles how dermatologists diagnose but also enjoys two additional benefits than conventional approaches. First, TATL provides additional information about the skin attribute regions regardless of their attributes, which can be helpful for dermatologists. Remarkably, such areas reveal variations and commonalities of relevant lesions, thereby reducing subjective errors in the evaluation process. This can be demonstrated by two examples in Figure \ref{fig:overlap-union-gt}. Here, attributes such as ``Negative Network” or ``Globules” can be challenging to identify in isolation. In contrast, the union of all attributes provided by TATL can correctly cover those areas. Second, adapting weights trained on skin attribute regions to a specific attribute can guide the network to pay attention to shared features across diverse attributes, thus strengthening trained systems to be more robust and stable. We will empirically demonstrate in detail these properties from Subsection \ref{sec:main-results} to Subsection \ref{sec:compute_bound}.
\section{Methodology}\label{Method}
We now detail our TATL framework and discuss the theoretical properties, which  sheds light on its {competitiveness}. We first provide an overview of our TATL framework in Section~\ref{sec:tatl}. Then, we discuss the pretext task with the Segment-Net and Attribute-Agnostic Segmenter in Section~\ref{sec:pretex}, and the downstream task with the Target Segmenter in Section~\ref{sec:downstream}. Then, we summarize the TATL framework and outline its algorithm in Section~\ref{sec:summ}. Lastly, we conclude this section with a theoretical insight in Section~\ref{sec:theory}.
\subsection{{Task Agnostic Transfer Learning for Skin lesion Attribute Detection}} \label{sec:tatl}
\begin{figure*}
	\begin{center}
	\includegraphics[width=1.0\textwidth]{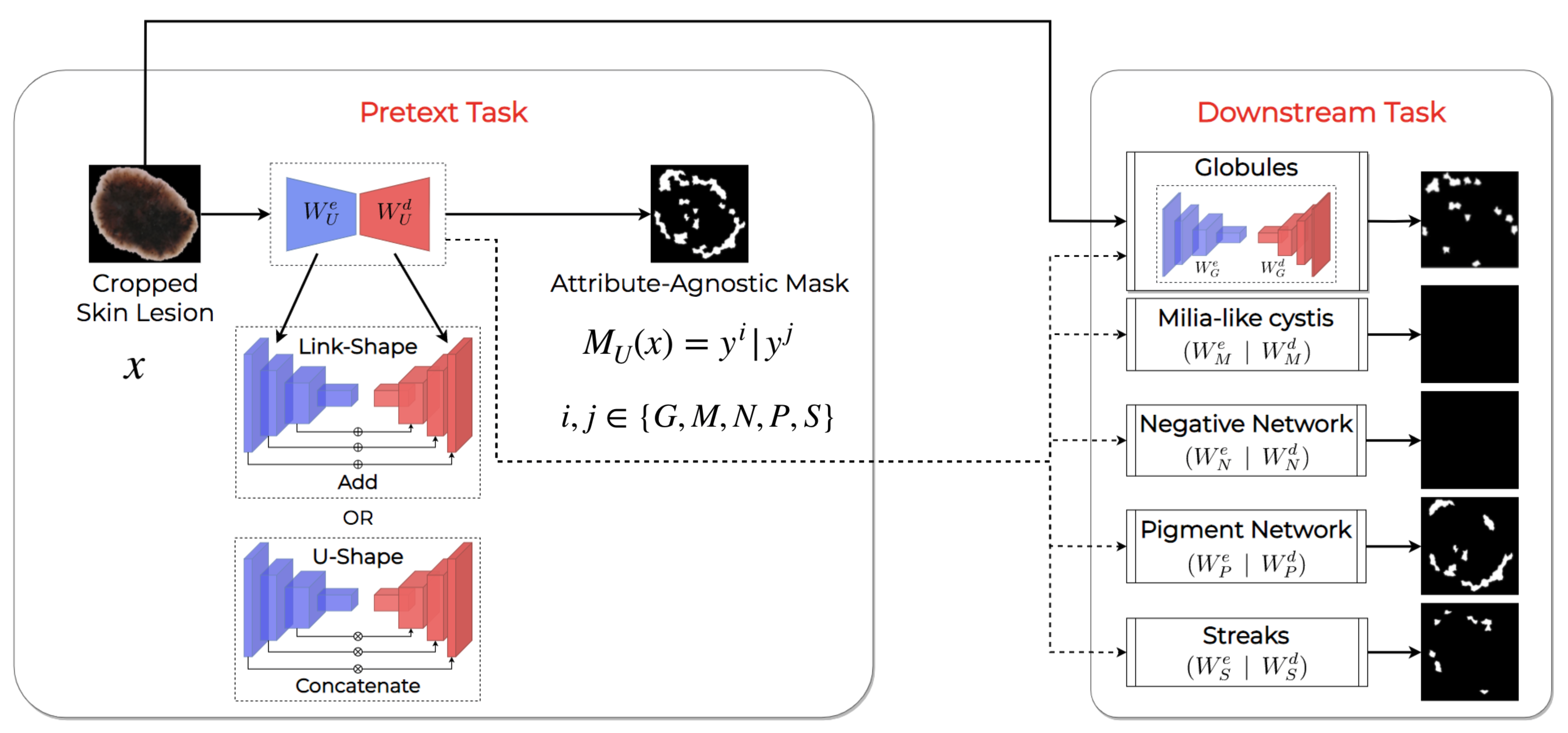}
	\caption{The training procedure of proposed TATL with two steps. Left: learning to recognize regions containing any attributes through attribute-agnostic mask (Pretext Task). Right: training parameters are transferred from Pretext Task to Downstream Task to segment each feature independently.}
	\label{fig:training}
    \end{center}
\end{figure*}
\paragraph{The Encoder - Decoder Architecture} The core component in our TATL framework is the encoder-decoder architecture, which takes an image as input and produces a binary mask as output. While we employ two kinds of encoder-decoder networks in our method, they share the same design as follows. The encoder part \textit{could be any feature extraction layers in} arbitrary architectures such as ResNet152~\citep{he2016deep} or EfficientNet \citep{tan2019efficientnet}. For feature extraction purposes, we thus discard non-linear rectification layers in these architectures. For the decoder path, we used layers to up-sample the encoder's features back to the original input's dimension. Particularly, to match the encoder's stages, the decoder consists of an up-sampling layer and a sequence of convolutional blocks where each block has two $3\times 3$ convolution filters with activation functions in between. Each stage in the decoder receives a feature map from its immediate preceding layer and a corresponding feature from the encoder's stage. The two inputs are combined by either the {\it adding} or {\it concatenating} operations, corresponding to the settings of LinkNet~\citep{chaurasia2017linknet} and U-Net~\citep{ronneberger2015u}.

\paragraph{The TATL Framework} Our TATL framework involves of {\it three} encoder-decoder networks. The first network, \textbf{Segment-Net}, segments and upscales the lesion regions in the original image. Then, the second network, \textbf{Attribute-Agnostic Segmenter}, takes the lesion regions as input and learns to segment the skin attribute regions, possibly including any of the five attributes of interests. Finally, for each attribute, a corresponding network, the \textbf{Target Segmenter}, is trained to segment that attribute's regions. Moreover, the Target Segmenter's decoder also rescales the final mask to match the original image's dimensions. Therefore, our TATL framework consists of {\it seven} networks in total, one Segment-Net, one Attribute-Agnostic Segmenter, and five Target Segmenters corresponding to five attributes. Each network uses either the {\it Link-shape} or {\it U-shape} architecture with the b0-EfficientNet \citep{tan2019efficientnet} as the main backbone due to its lightweight property compared to other architectures (Table \ref{tab:parameter}).

We now introduce some notations to detail our framework. We denote  $\{f_{Seg}(., W_{seg}),\, f_{U}(., W_{U}),\, f_i(., W_{i})\}$ as the corresponding networks of the Segment-Net, the Attribute-Agnostic Segmenter, and the Target-Segmenter for each {attribute} $i \in Y = \{G, M, N, P, S\}$ in Stage 3 respectively (Figure \ref{fig:prediction-method}). We also use $W_m =\{W_{m}^{e}, \, W_m^{d}\},\ m \in \{Seg, U, G, M, N, P, S\}$ as the total parameters of networks $f_m$, where $\{W_m^{e}\}$, and $\{W_{m}^{d}\}$ represent weights of encoder and decoder layers. To train the Segment-Net, we denote $D_{seg} = \{x, M_{seg}\}$ as the set of images and their corresponding lesion region masks. 

\subsection{Pretext Task with Segment-Net and Attribute-Agnostic Segmenter} \label{sec:pretex}

Here, we refer to the TATL \textit{Pretext Task} as the problem of recognizing regions containing any attributes. The pretext training scheme consists of two stages: (i) cropping skin lesion with the Segment-Net; and (ii) segmenting the attribute-agnostic mask with Attribute-Agnostic Segmenter. In the following, we describe the training procedure with the corresponding components in detail.

\subsubsection{Segment-Net}
In Stage 1, we use the Segment-Net on the lesion dataset $D_{seg}$ to eliminate extraneous skin-based regions and only keep the lesion regions. Especially, given an original image, we re-scale it to a size of $386 \times 512 $ and pass it into the Segment-Net. A bounding box with an offset value of $40$ pixels in four directions is used to crop the Segment-Net's output so that errors in the segment step are not propagated in the later stages. Once this bounding box has been created, it is scaled to match the resolution of the input image and utilized as a new input for the next stage.

\subsubsection{Attribute-Agnostic Segmenter}

The second stage focuses on training the Attribute-Agnostic Segmenter. We first define the {\it Attribute-Agnostic} region as a region that contains at least one of the attributes in $Y$. From this, we define an intermediate dataset of the Attribute-Agnostic as $D_U = \{x, M_U\}$, where $M_U$ is the binary mask corresponding to an image whose value is $1$ whenever a pixel is an attribute from $Y$ (Pretext Task). Note that given an image $x$ and a set of attributes masks, $M_U$ is the {\it union} of all the masks and can be easily constructed by performing bitwise OR operator as:

\begin{equation}
    M_U(x) \triangleq  y^{(1)} \BitOr y^{(2)} \BitOr \ldots \BitOr y^{(|Y|)},
\end{equation}

where $\BitOr$ denotes the bitwise OR operator.
The dataset $D_U$ is used to train the Attribute-Agnostic Segmenter such that it can detect the skin attribute regions belonging to any of the attributes. 
It is important to note that $D_U$ contains masks covering all attributes, thus ameliorating the negative effect of training data scarcity, especially for minor classes.

\subsection{Downstream Task with Target Segmenters} \label{sec:downstream}
Given $W_{U}$ learned from the Attribute-Agnostic Segmenter, we can proceed to Stage 3 and train the segmenter for each of the attributes (\textit{Downstream Task}). Different from previous approaches, we initialize the Target-Segmenter's parameters from the Attribute-Agnostic Segmenter's parameters as: $W_i^{e} \leftarrow W_{U}^{e}$ and $W_{i}^{d} \leftarrow W_{U}^{d}$ for each type $i$- attribute. 
Lastly, a set of \textit{ Target-Segmenters} is trained to segment the attributes.

Having a dedicated network for each attribute is advantageous in alleviating the imbalance training data problem. Moreover, we explore two strategies in training the \textit{Target-Segmenters}, which corresponds to allowing knowledge sharing across attributes or not. First, we {\it freeze} all the encoders (\textbf{TATL-Freeze}) to allow feature sharing across attributes because the encoder is initialized from the Attribute-Agnostic Segmenter. Second, we allow both the encoder and decoder to be updated (\textbf{TATL-Non Freeze}), which allows each Target-Segmenter to adapt to their dedicated attribute. We summarize  these Pretext (with Attribute-Agnostic Segmenter) and Downstream Tasks  in Algorithm \ref{alg:trans} and Figure \ref{fig:training}. Ablation studies for freeze encoder layers are also discussed in Table \ref{tab:result1718} of Subsection~\ref{sec:tatl-various-architectures}.

\begin{algorithm}[t]
\SetAlgoLined
\DontPrintSemicolon

\KwInput{
Pre-trained ImageNet from employed backbone $W_{\mathrm{ImgNet}} = \{W_{\mathrm{ImgNet}}^{e}, \, W_{\mathrm{ImgNet}}^{d}\}$ \\
The attribute dataset:   ${\mathcal{D}= \{(x_1, \{y_1^{(i)}\}_{i=1}^{|Y|} ) \ldots, (x_n, \{y_n^{(i)}\}_{i=1}^{|Y|})\}}$}
\KwOutput{Trained weights $W_i = \{W_i^{e}, W_i^{d}\},\, i \in Y = \{G, M, N, P, S\}$} 
\tcp{Create the attribute-agnostic masks}
\For{\textup{each image}  $\boldsymbol{x} \in \mathcal{D}$}{
    $\boldsymbol{M_U}[x] = y^{(1)} \BitOr y^{(2)} \BitOr \ldots \BitOr y^{(|Y|)}$\;
}
\tcp{Learning \textit{Attribute-Agnostic Segmenter}}
    \textbf{Initialise: }$W^e_U \leftarrow W^e_{\mathrm{ImgNet}}$ and $W^d_U \leftarrow W^d_{\mathrm{ImgNet}}$\;
    
\For{\textup{minibatch} $\{\boldsymbol{x_k}, \, \boldsymbol{y_k}\}$ \textup{where}  $\boldsymbol{x_k} \in \mathcal{D}$ , $\boldsymbol{y_k} \in \boldsymbol{M_U}$}{
    Minimize $\widehat{f}_U(\{x_k,\,y_k\}|W_U)$ using Eq. (\ref{eq:1})\\
Update $W^e_{U}$, $W^d_{U}$
}

\tcp{Learning \textit{Target-Segmenter} for each attribute}
\For{\textup{attribute} $i \in Y$}{
    \textbf{Initialise: }$W^e_i \leftarrow W^e_U$ and $W^d_i \leftarrow W^d_U$\;
    \For{\textup{minibatch} $\{\boldsymbol{x_k}, \, \boldsymbol{y_{k}^{i}}\}$ \textup{where}  $\boldsymbol{x_k}, \boldsymbol{y_{k}^{i}} \in \mathcal{D}$}{
            Minimize $\widehat{f}_i(\{x_k,\,y_{k}^{i}\}|W_i)$ using Eq. (\ref{eq:1})\\
        Update $W^e_{i}$, $W^d_{i}$
    }
}
\textbf{return} $W^e_i$ and $W^d_i$ where $i \in Y$ 
\caption{The TATL Algorithm}
\label{alg:trans}
\end{algorithm}

\subsection{Summary of TATL} \label{sec:summ}
We summarize the training and inference pipeline of TATL as follows,
\paragraph{Training Step}
Given a training dataset, TATL training performs the following steps,
\begin{enumerate}
    \item Training the Segment-Net using dataset $D_{seg} = \{x, M_{seg}\}$.
    \item Training Attribute-Agnostic Segmenter and  
    Target-Segmenters using Algorithm \ref{alg:trans} and dataset \newline
    ${\mathcal{D}= \{(x_1, \{y_1^{(i)}\}_{i=1}^{|Y|} ) \ldots, (x_n, \{y_n^{(i)}\}_{i=1}^{|Y|})\}}$
\end{enumerate}
Note that we only apply a U-Shape encoder-decoder architecture in the first step, while we use both U-shape and Link-shape connections in the second step. 

\paragraph{Inference Step}
Given an input image, it first will be segmented by the Segment-Net and then fed into five different Target-Segmenters to segment five kinds of skin features. To compare with other competitors, we selected the b0-EfficientNet \citep{tan2019efficientnet} as the network backbone and used either the \textit{adding} (Link-shape) and \textit{concatenating} (U-shape) operations to correlate feature maps obtained from encoder and decoder layers. The final binary map for each skin attribute is produced by averaging the probability estimates from these two network architectures. Note that our pipeline prediction requires a focused lesion image generated by the Segment-Net, which might be influenced by the employed segmentation method. Fortunately, TATL only uses a bounding box with a specific offset around predicted lesion regions, thus making the following steps less susceptible to the resulting segmentation step. We present an experiment in Table \ref{tab:segmentation_impact} to validate this property.

\subsection{Theoretical Insights} \label{sec:theory}
\label{sub:theoretical}
This section provides theoretical insights to justify our approach using recent results from data-dependent stability of optimization schemes. 
Particularly, we 
investigate the model's generalization to a target domain based on its initialization weights and show that initializing in a TATL fashion gives a tighter generalization error bound than ImageNet initialization via Proposision~\ref{pro:proposition}.

First, we introduce used notations, which are briefly summarized in Table~\ref{tab:notations}. 
\begin{table*}
\centering
    \caption{Summary of used notations.}
    \vspace{0.1in}
\scalebox{1.0}{
    \setlength\tabcolsep{3pt}
	\begin{tabular}{|c|c|}
		\hline
		\textbf{Notations} & \textbf{Definitions}\\\hline
		X, Y & The input and output spaces \\
		S & the training set\\
		$Z$ & the joint input-output space ($Z=X\times Y$)\\
		$\{\alpha_t\}_{t=1}^T$ & step sizes\\
		$I = \{j_t\}_{t=1}^T$ & random indices\\
		$D$ & the data-generating distribution \\
		$H$ & the hypothesis space \\
		$w$ & a hypothesis \\
		$A_S$ & learning algorithm given the training data $S$ \\
		$\epsilon(D,w_1)$ & stability function of $D$ and $w_1$
		\\\hline
    \end{tabular}
}
\label{tab:notations}
\end{table*}
We consider the supervised training problem with $X \subset \mathbb{R}^n$ as the input space and $Y \subset \mathbb{R}$ as the output space.
We also assume that training and testing instances are sampled i.i.d. from a probability distribution $D$ over $Z = X\times Y$. Also, we denote the training set $S = \{z_{i}\}_{i=1}^{m} \stackrel{iid}{\sim}  D$, where a training sample $z_i$ consists of an input image $x \in X$ and its corresponding label $y \in Y$. We express $H=\{w_j \}$ as the hypothesis space where $w_j \in \mathbb{R}^d$ denotes a hypothesis (model) with dimension $d$ that maps an input instance $x$ to its corresponding label $y$.
Lastly, we define a map $A_S: S \rightarrow H$ as a learning algorithm that returns a hypothesis given a training data set $S$.

\citet{kuzborskij2018data} established a data-dependent aspect of algorithm stability for Stochastic Gradient Descent (SGD) given a training set $S = \{z_i\}_{i=1}^{m} \stackrel{iid}{\sim} D$, step sizes $\{\alpha_{t}\}_{t=1}^{T}$, random indices $I=\{j_{t}\}_{t=1}^{T}$, and an initialization weight $w_1$, which is sequentially updated as:
\begin{equation}
    w_{t+1} = w_{t} - \alpha_{t}\nabla \ell(w_{t}, z_{{j}_{t}}).
\end{equation}

Here, $\ell(w, z)$ is a loss function, which measures the difference between predicted values and true values with parameters $w \in H$ on an sample $z$. We indicate $(D, w_1)$ as the data-generating distribution and the initialization point $w_1$ of SGD, $\epsilon(D, w_1)$ as a stability function of $D$ and $w_1$. 

To characterize a randomized learning algorithm $A$, we define its ``On-Average stability''.
\begin{defi}
(On-Average stability). A randomized algorithm $A$ is $\epsilon(D, w_1)$-on-average stable if
\[
\mathrm{sup}_{i\in[m]}\Big\{\mathbb{E}_{S}\mathbb{E}_{S,z}\big[f(A_{S}, z) - \ell(A_{S^{(i)}}, z)\big]\Big\} \leqslant \epsilon(D, w_1),
\]
where $S \stackrel{iid}{\sim} D$ and $S^{(i)}$ is S copy with i-th example replaced by $z \stackrel{iid}{\sim} D$.
\end{defi}
We now have the following theorem \citep{kuzborskij2018data}:
\begin{thr}
\vspace{-0.1in}
\label{theorem1}
Let $A$ be $\epsilon(\theta)$ on average stable, then
\[
    \mathbb{E}_{S}\mathbb{E}_{A}[R(A_{s}) - \hat{R}_{S}(A_s)] \leqslant  \epsilon(D,w_1),
    \]
where $R(A_{s}),\, \hat{R}_{S}(A_s)$ are risk and empirical risk of $A_s$ respectively, defined by:
\[
R(A_{s}) = \mathbb{E}_{z \sim D}\big[\ell(A_{s}, z)\big];\  \hat{R}_{S}(A_s) = \frac{1}{m} \sum_{i=1}^{m}\, \ell(A_{s}, z_{i}).
\]
\end{thr}
In other words, the generalization of a learning algorithm on unseen data drawn from the same distribution is controlled by its $\epsilon(D,w_1)$-on average stable, which depends on initialized weights $w_1$. In the following, we will examine the model's generalization performance through the lens of its training algorithm's stability.

In the transfer learning setting, Theorem~\ref{theorem1} provides a tool to understand the model's generalization on the target domain, given that it is initialized from one of the pre-trained models on a set of source domains (source tasks). Specifically, we suppose that the target task is characterized by a joint probability distribution $D^{\mathrm{tgt}}$ and assume that a set of source hypotheses $\{w_{k}^{\mathrm{src}}\} \subset H, k \in K$ trained on $K$ different source tasks. In this paper, we consider two distinct source cases with $K = \{\mathrm{ImgNet}, \mathrm{TATL}\}$ where ``$\mathrm{ImageNet}$'' refers to weights trained on ImageNet and ``$\mathrm{TATL}$'' is our approach of learning the attribute-agnostic mask. 
Now we are ready to analyze the generalization bound of TATL versus ImageNet initialization strategies by utilizing the results in \citet{kuzborskij2018data}.

\begin{pro}
\label{pro:proposition}
Given a non-convex loss function $\ell$ and assume that $\ell(., z) \in [0,1]$ has a $p$-Lipschitz Hessian, $\beta$-smooth  and step sizes of a form $\alpha_{t} = \frac{c}{t}$ satisfy $c \leqslant \min(\frac{1}{\beta}, \, \frac{1}{4(2\beta \mathrm{ln}(T))^{2}})$, then with high probability, the $\epsilon(D^{\mathrm{tgt}}, w_k^{src})$ of SGD scheme satisfies:
\begin{equation} \label{eqn:prop1}
\underset{k \in K}{\mathrm{min}}\, \epsilon(D^{\mathrm{tgt}}, w_{k}^{\mathrm{src}}) \leqslant
 \underset{k \in K}{\mathrm{min}}\, \mathcal{O}\bigg( \Big(1+ \frac{1}{c\hat{\gamma}_{k}^{-}}
\Big)\hat{R}_{S}(w_{k}^{\mathrm{src}})^{\frac{c\hat{\gamma}_{k}^{+}}{1+c\hat{\gamma}_{k}^{+}}}. \frac{\sqrt{\log (|K|)}}{m^{\frac{1}{1 + c\hat{\gamma}_{k}^{+}}}}
\bigg),
\end{equation}
where \begin{align}
    \hat{\gamma}_{k}^{\pm} &= \hat{\gamma}_{k} \pm \frac{1}{\sqrt[4]{m}}, \\
    \hat{\gamma}_{k} &= \frac{1}{m} \sum_{i=1}^{m} ||\nabla^{2}\ell(w_{k}^{\mathrm{src}}, z_{i})||_{2} + \sqrt{\hat{R}_{S}(w_{k}^{\mathrm{src}})},
\end{align}
\end{pro}
with $||.||_2$ is a spectral norm. Intuitively, Theorem~\ref{theorem1} and Proposition~\ref{pro:proposition} suggest that an initialization's generalization error depends on {\it two} factors: (i) how well it performs on the target domain without any training, which is characterized by $\hat{R}_S$; and (ii) the loss function's curvature around this initialization, which is characterized by empirical $||\nabla^{2}\ell(w_{k}^{\mathrm{src}}, z)||_{2}$ over $m$ training samples, denoted as $\hat{\gamma}$. 
This result provides an intuitive explanation of why TATL provides a more favorable initialization than the traditional ImageNet pre-trained models. Notably, we will explain why TATL, which initializes the Target-Segmenter from the Attribute-Agnostic Segmenter, can perform better than initializing the segmenter from ImageNet pre-trained models. 

Pre-trained ImageNet models are unlikely to perform well on medical images due to the vast diversity between the two domains. Therefore, such models often have a higher empirical error on the target domain $\hat{R}_S$ and usually lie in {\it high curvature} regions $\hat{\gamma}$. On the other hand, TATL uses an initialization from the Attribute-Agnostic Segmenter, which is pre-trained on self-generated data of the target task. Note that the Attribute-Agnostic Segmenter can detect {\it any} of the attributes, and therefore enjoy {\it lower} empirical risk $\hat{R}_S$ compared to ImageNet models. Moreover, due to its construction, the Attribute-Agnostic Segmenter's parameter likely lies in a region close to the local minimum of each attribute detector, which enjoys {\it lower curvature} $\hat{\gamma}$. Consequently, TATL exploits the target task's knowledge to form an initialization with a high probability of attaining lower empirical error and curvature, which translates to a tighter generalization error bound than initializing from pre-trained ImageNet models. We empirically verify this hypothesis by comparing the bound's values in Eq.~(\ref{eqn:prop1}) of different initialization strategies in Figure~\ref{fig:graph} of Section~\ref{sec:compute_bound}.

\section{Experiments and Results}
\subsection{Dataset}
We conduct experiments on two well-known datasets for skin attributes detection: the ISIC 2017\footnote{\url{https://challenge.isic-archive.com/landing/2017}} and 2018\footnote{\url{https://challenge2018.isic-archive.com/}} Task 2 datasets. Table~\ref{tab:descr} provides a summary of the two datasets. It is worth noting that the ISIC 2017 dataset only contains {\it four} classes: Streaks, Negative Network, Milia, and Pigment Network, while the ISIC 2018 introduces a new class of Globules. Moreover, both datasets exhibit high data imbalance among the attributes. For example, in the ISIC 2018 dataset, the class ``Streaks'' only appears in 3.86\% of the training data while ``Pigment Network'' is observed in 58.67\% of the training data.

\begin{table*}[!hbt]
\centering
\caption{Distribution of training images in the ISIC 2017 and 2018 Task 2 datasets.}
\vspace{0.1in}
\label{tab:descr}
\scalebox{1.0}{
\setlength\tabcolsep{3.5pt}
\begin{tabular}{|c|c|c|c|c|c|c|c|}
\hline
\multicolumn{1}{|l|}{}                & \textbf{Class} & \textbf{Streaks} & \textbf{\begin{tabular}[c]{@{}c@{}}Negative\\ Network\end{tabular}} & \textbf{Globules} & \textbf{Milia} & \textbf{\begin{tabular}[c]{@{}c@{}}Pigment\\ Network\end{tabular}} & \textbf{Total} \\ \hline \hline
\multirow{2}{*}{\textbf{ISIC - 2017}} & Number         & 113              & 122     & NA                & 475            & 1119   & 1416           \\
 & Rate (\%)      & 7.98             & 8.62     & NA                & 33.55         & 79.03  & 100\%          \\ \hline
\multirow{2}{*}{\textbf{ISIC - 2018}} & Number         & 100              & 190     & 602               & 681            & 1522   & 2594           \\
 & Rate (\%)      & 3.86              & 7.32     & 23.21              & 26.25           & 58.67   & 100\%          \\ \hline
\end{tabular}}
\end{table*}

\subsection{Experimental Settings}
We implemented all experiments using the Pytorch framework~\citep{paszke2019pytorch} on 4 NVIDIA TITAN RTX GPUs. All images were pre-processed by centering and normalizing the pixel density per channel. Besides, we also re-scale all images to the resolution of $386 \times 512$ in training steps and transform final predictions to a size of $256 \times 256$ in the evaluation step following the standard of the ISIC challenge. We used the SGD optimizer~\citep{goodfellow2016deep} with an initial learning rate of $0.01$ and momentum of $0.9$ to be consistent with the theory presented in Section \ref{sub:theoretical}.
For TATL, we obtained the Segment-Net by training a b0-EfficientNet backbone with {\it U-shape} on both the ISIC 2018 and ISIC 2017 Task 1 using the loss function in Eq.~(\ref{eq:1}).
Given the segmentation results, we defined a bounding box around the masks with an offset of 40 pixels in four directions to mitigate the segmentation errors, {before feeding them to the Attribute-Agnostic Segmenter and the Target-Segmenters}. The Attribute-Agnostic Segmenter and the Target-Segmenters were then trained for 40 epochs with early-stopping after 10 epochs. Finally, we measure our performance and compare it with other baselines using the five-fold cross-validation method and report the average values on Dice and Jaccard coefficients as \citet{koohbanani2018leveraging}.

\subsection{Comparison Against Other Approaches} \label{sec:main-results}
We compare our method against the winner of ISIC 2017 \citep{kawahara2018fully} and  ISIC 2018 \citep{koohbanani2018leveraging} and report the Dice and Jaccard index in Table~\ref{tab:our_result2017} and Table~\ref{tab:our_result2018} respectively. Here the results of the 1-st method in ISIC 2018 are extracted from their original paper \citep{koohbanani2018leveraging} while we use the source code published by \citep{kawahara2018fully} and run-again experiments using our setting with the five-fold cross-validation method.
\begin{table*}[!hbt]
\centering
\caption{Jaccard and Dice metrics on the ISIC2018 challenge.  Blue and Red colors are the best results in Jaccard and Dice for each attribute, respectively. Stage 1: segmenting the lesion region, Stage 2: training the Attribute-Agnostic Segmenter, Stage 3: training the Target-Segmenters.}

\vspace{0.1in}
\label{tab:our_result2018}
    \setlength\tabcolsep{3pt}
    \centerline{
	\begin{tabular}{|c|c|c|c|c|c|c|c|c|c|c|c|c|}
		\hline
		\textbf{Method}    & \multicolumn{2}{c|}{  ISIC2018 1st}   & \multicolumn{2}{c|}{ \begin{tabular}[c]{@{}c@{}}TATL \\\textbf{Stage 3} \end{tabular}}& \multicolumn{2}{c|}{\begin{tabular}[c]{@{}c@{}} TATL \\ \textbf{Stage 2, 3}\end{tabular}} & \multicolumn{2}{c|}{\begin{tabular}[c]{@{}c@{}}  TATL \\ \textbf{Stage 1, 3}\end{tabular}} & \multicolumn{2}{c|}{\begin{tabular}[c]{@{}c@{}}  TATL \\ \textbf{Stage 1, 2, 3}\end{tabular}} &
		\multicolumn{2}{c|}{\begin{tabular}[c]{@{}c@{}}  TATL w/o ImgNet\\ \textbf{Stage 1, 2, 3} \end{tabular}} 
		\\ \cline{2-13} 
		&  Jaccard  & Dice&  Jaccard & Dice   &  Jaccard  & Dice &  Jaccard   & Dice&  Jaccard   & Dice & Jaccard   & Dice \\ \hline\hline
		\textbf{Pigment Net.} & 0.563  & 0.720 &0.532  &0.691   &0.580  &0.721  &0.542 &0.701 &\textcolor{blue}{0.584} &\textcolor{red}{0.730 } &0.565 & 0.721\\
		\textbf{Globules} & 0.341 & 0.508 &0.308 &0.471   &0.368   &\textcolor{black}{0.546}  &0.332 &0.467 &\textcolor{blue}{0.379} &\textcolor{red}{0.552}  &0.359 & 0.528\\
		\textbf{Milia-like cysts} & \textcolor{black}{0.171} & \textcolor{red}{0.289} &0.141  &0.252   &0.161   &0.268 &0.142 &0.264 &\textcolor{blue}{0.172} &\textcolor{black}{0.288} & 0.161 & 0.277  \\
		\textbf{Negative Net.}& 0.228  & 0.371   &0.149  &0.260  &\textcolor{black}{0.269}   &\textcolor{black}{0.403}  &0.194 &0.348 &\textcolor{blue}{0.283} & \textcolor{red}{0.438} & 0.280 &  0.437\\
		\textbf{Streaks}  & 0.156  & 0.270   &0.139  &0.241   & 0.254  &\textcolor{black}{0.394}  &0.135 &0.224 &0.254 &0.401 & \textcolor{blue}{0.263} & \textcolor{red}{0.416}  \\ \hline
		\textbf{Average}  & 0.292  & 0.432   &0.254  &0.383   &0.326   &0.466  &0.269 &0.401  &\textcolor{blue}{0.334} & \textcolor{red}{0.482} & 0.326 & 0.476
		\\\hline
    \end{tabular}
    }
\end{table*}
\begin{table*}[!hbt]
\centering
\caption{Jaccard and Dice metrics on the ISIC2017 challenge.  Blue and Red colors are best results in Jaccard and Dice for each attribute, respectively. Stage 1: segmenting the lesion region, Stage 2: training the Attribute-Agnostic Segmenter, Stage 3: training the Target-Segmenters.}
\vspace{0.1in}
\label{tab:our_result2017}

    \setlength\tabcolsep{3pt}
    \centerline{
	\begin{tabular}{|c|c|c|c|c|c|c|c|c|c|c|c|c|}
		\hline
		\textbf{Method}    & \multicolumn{2}{c|}{  ISIC2017 1st}   & \multicolumn{2}{c|}{ \begin{tabular}[c]{@{}c@{}}TATL \\\textbf{Stage 3} \end{tabular}}& \multicolumn{2}{c|}{\begin{tabular}[c]{@{}c@{}} TATL \\ \textbf{Stage 2, 3}\end{tabular}} & \multicolumn{2}{c|}{\begin{tabular}[c]{@{}c@{}}  TATL \\ \textbf{Stage 1, 3}\end{tabular}} & \multicolumn{2}{c|}{\begin{tabular}[c]{@{}c@{}}  TATL \\ \textbf{Stage 1, 2, 3}\end{tabular}} &
		\multicolumn{2}{c|}{\begin{tabular}[c]{@{}c@{}}  TATL w/o ImgNet\\ \textbf{Stage 1, 2, 3} \end{tabular}} 
		\\ \cline{2-13} 
		&  Jaccard  & Dice&  Jaccard & Dice   &  Jaccard  & Dice &  Jaccard   & Dice&  Jaccard   & Dice & Jaccard   & Dice \\ \hline\hline
		\textbf{Pigment Net.} & 0.389 & 0.556  &0.426  &0.597  &0.499 &0.667  &0.497 & 0.665 & \textcolor{blue}{0.516}  &\textcolor{red}{0.681} & 0.473 & 0.639 \\
		\textbf{Milia-like cysts} & \textcolor{blue}{0.119} & \textcolor{red}{0.215} &0.072  &0.127  &0.091  &0.157 &0.101 &0.172 &0.108 &0.188 & 0.092 & 0.168\\
		\textbf{Negative Net.}& 0.201  & 0.333   &0.126  &0.218 &0.191   & 0.310 &0.147 &0.251 &0.213 &0.334 & \textcolor{blue}{0.234} &  \textcolor{red}{0.380}\\
		\textbf{Streaks}  & 0.192   & 0.321  &0.139 &0.233   &0.203  &0.336 &0.139  &0.232 &\textcolor{blue}{0.215} &\textcolor{red}{0.346} & 0.209 & 0.345 \\ \hline
		\textbf{Average}  & 0.225  &  0.356  &0.191 & 0.294 &0.246  &0.367 &0.221 &0.330  &\textcolor{blue}{0.263} & \textcolor{red}{0.387} & 0.252 & 0.383
		\\\hline
    \end{tabular}
    }
\end{table*}

\subsubsection{TATL Variations}
Due to the high competitiveness of skincare challenges, we utilize both U-shape and Link-shape architectures with b0-EfficientNet \citep{tan2019efficientnet} as the backbone for the Attribute-Agnostic Segmenter and Target-Segmenters, then taking the average probability predictions. For a comprehensive comparison, we include {\it five} variants of four TATL corresponding to removing components in the TATL framework: 
\begin{itemize}
    \item the vanilla encoder-decoder architecture but without the Segment-Net and the Agnostic-Attribute Segmenter (\textit{TATL \textbf{Stage} 3)}; 
    \item a variant that performs the second and last stage of our TATL: train first the Attribute-Agnostic segmenter on the original images and then a set of Target-Segmenters (\textit{TATL \textbf{Stage 2, 3}}); 
    \item a variant that performs the first and last stage of our TATL: segment the lesion regions and then the attributes (\textit{TATL \textbf{Stage 1, 3}}); 
    \item our full TATL framework that performs all three stages (\textit{TATL \textbf{Stage 1, 2, 3}});
    \item our full TATL framework initialized from scratch (\textit{TATL w/o ImgNet \textbf{Stage 1, 2, 3}}).
\end{itemize}

\subsubsection{Overall performance}
Our TATL shows competitive performances against other approaches on both benchmarks and metrics. Notably, our method presents substantial improvements over other baselines on attributes with the least amount of training data. For example, in Table \ref{tab:our_result2018}, \textbf{\textit{TATL  Stage 1, 2, 3}} for Negative Network achieved the Jaccard index of 0.283, which is 5.5\% higher compared to ISIC 2018 winner (0.228) \citep{koohbanani2018leveraging}. Also, this TATL setting for the Streak feature has the Dice index of 0.401, which is 13.1\% higher compared to ISIC 2018 winner (0.270) \citep{koohbanani2018leveraging}. 

\subsubsection{{Ablation Study of TATL's Sub-stages}}
This experiment aims to investigate the contribution of the first and second stages to the final performance, given that Stage 3 is always enabled for supervised training. 
Table \ref{tab:our_result2018} and Table~\ref{tab:our_result2017} demonstrate that enabling Stage 2, i.e., \textit{TATL \textbf{Stage 2, 3}} results in a significant improvement in most skin attributes compared to using Segment-Net (\textit{TATL \textbf{Stage 1, 3}}). For example, the Average Dice score in Table \ref{tab:our_result2018} of \textit{TATL \textbf{Stage 3}} increased from 0.383 to 0.466 with \textit{TATL \textbf{Stage 2, 3}} while only attaining at 0.401 with \textit{TATL \textbf{Stage 1, 3}}. This result thus emphasizes the critical role of learning the Attribute-Agnostic Segmenter in our framework.

In addition, progressively adding Stage 1 to the Stage 2 and 3 model further improves the results. For example, in Table~\ref{tab:our_result2017}, when using a pre-trained ImageNet, adding Stage 1 increased the Averaged Dice score from 0.367 to 0.387. Overall, the findings support our hypothesis that all three phases contribute to TATL's competitive performance.

\subsubsection{The Influence of using Pre-trained ImageNet in TATL}
We investigate the advantages of employing pre-trained models in TATL (Algorithm 1) by examining the \textit{TATL w/o {ImgNet} {Stage 1, 2, 3}}. The outcomes suggest that employing pre-trained models in TATL favours classes with more training data. Remarkably, for ISIC 2017's Pigment Network, the attribute with the most training samples, pre-trained models improves the Jaccard index from 0.473 to 0.516 (9.09\% relative improvement). On the other hand, the contributions of pre-trained models on attributes with limited training samples such as Streaks and Negative Network are much less significant, e.g., ISIC 2017's Negative Network Jaccard index increased from 0.209 to 0.215 (2.87\% relative improvement). In general, we conclude that the improvements observed in minor classes come from our TATL framework rather than the pre-trained ImageNet;  however, the TATL version with pre-trained ImageNet, on average, outperforms the TATL version using random weights.

\begin{table*}[!hbt]
\centering
\caption{The average performance of the skin feature segmentation step with various lesion inputs generated by various networks and offset settings. Dice score is used to measure  accuracy.}
\vspace{0.1in}
\begin{tabular}{|c|c|c|c|c|c|}
\hline
\multirow{2}{*}{\textbf{Method}} & \multirow{2}{*}{\textbf{\begin{tabular}[c]{@{}c@{}}Performance of \\ Segmented Skin Lesion\end{tabular}}} & \multicolumn{4}{c|}{\textbf{Performance of Segmented Skin Features}} \\ \cline{3-6} 
                                 &                                                                                                           & Offset = 0     & Offset = 20     & Offset = 40     & Offset = 60     \\ \hline \hline
U-Net                            & 0.783                                                                                                     & 0.358          & 0.362           & 0.387           & 0.376           \\ \hline
SegNet                           & 0.824                                                                                                     & 0.365          & 0.371           & 0.387           & 0.376           \\ \hline
Mask-RCNN                        & 0.876                                                                                                     & 0.371          & 0.372           & 0.387           & 0.376           \\ \hline
\end{tabular}
\label{tab:segmentation_impact}
\end{table*}

\subsubsection{The Influence of Cropping Lesion Segmentation on the Final Result}
{Our inference step requires the segmented lesion region to eliminate less relevant parts for the later phase. This task is handled by the Segment-Net (Stage 1). The lesion regions then are cropped by a bounding box with an offset value in four directions to generate input for the next step. In Table \ref{tab:our_result2018} and Table \ref{tab:our_result2017}, we presented the ablation study for  Stage 1 derived from \textit{U-Shape} using a b0-EfficientNet backbone with an offset of $40$ pixels. We now investigate how much the errors in Stage 1 can propagate to the final predictions by varying \textit{segmentation methods} and \textit{offset values}.} 

We conducted tests on ISIC 2017 in which four different models were trained to segment four skin features using the same configuration in Stage 2 and 3 but will take distinct inputs in Stage 1 created by various networks as U-Net \citep{ronneberger2015u}, SegNet \citep{badrinarayanan2017segnet}, and Mask-RCNN \citep{he2017mask}. In addition, we changed offset values ranging from 0 to 60 pixels with a 20-pixel step. Table \ref{tab:segmentation_impact} depicts the results of various approaches when changing these factors, in which we used the Dice score to compute accuracy for all experiments.

We observe that applying an offset $0$ pixel reduces the accuracy of the subsequent step because segmentation errors lead to the loss of some essential information, particularly at the image's border locations. It also explains why improved performance in the feature segmentation stage results from higher accuracy in the lesion segmentation step. When increasing the magnitude of offset to $20$ pixels, all methods are improved; for instance, the U-Net case rises from 0.358 to 0.362. Moreover, with the offset $40$ and $60$ pixels, margins between baselines are no longer available, and the performance of two later cases is better than offset $20$ pixels. Table \ref{tab:segmentation_impact} also  presents a trade-off in selecting large offset values. In particular, a considerable value of $60$ pixels can reduce efficiency compared to 40 pixels because the image may involve more unrelated data. {In summary, we conclude that adding an optimized offset value of 40 pixels around segmented lesion areas allows our inference step to be stable despite lesion segmentation perturbations.}

\subsubsection{Performance of Each Network Backbone}
{
This experiment examines the contribution of each component network to the overall performance of TATL.
In particular, we evaluate our method’s performance using just U-shape or Link-shape based on the b0-EfficientNet backbone and compare them to networks used in the ISIC-2018-1st: ResNet-151, Resnet-v2, and DenseNet-169.  Furthermore, we also include the b0-EfficientNet performance in the ISIC-2018 challenge for overall comparison. Table \ref{tab:backbone} highlights the acquired data, with blue and red representing the best Jaccard and Dice scores, respectively. In this table, our two variants, labeled as \textit{U-Eff(TATL)} and \textit{L-Eff(TATL)}, are the results of employing the U-shape and the Link-shape, respectively.}

\begin{table*}[!hbt]
\centering
\caption{The performance of ISIC2018-1st using a single backbone compared to our framework on the ISIC 2018. Blue and Red colors are the best values in Jaccard and Dice metrics.}
\vspace{0.1in}
\label{tab:backbone}
		\scalebox{1.0}{	
		\hspace{-0.2in}
		\setlength\tabcolsep{3pt}
		\begin{tabular}{|c|c|c|c|c|c|c|c|c|c|c|c|c|}
				\hline
				\textbf{Method}    & \multicolumn{2}{c|}{ \small \textbf{Pigment Net.}}   & \multicolumn{2}{c|}{\small  \textbf{Globules}}& \multicolumn{2}{c|}{\begin{tabular}[c]{@{}c@{}} \small  \textbf{Milia-like}\\ \textbf{cysts}\end{tabular}} & \multicolumn{2}{c|}{\small  \textbf{Negative Net.}} & \multicolumn{2}{c|}{\small  \textbf{Streaks}} & \multicolumn{2}{c|}{ \small  \textbf{Average}}\\ \cline{2-13} 
				& \small  Jaccard  &\small  Dice& \small  Jaccard & \small  Dice   & \small  Jaccard  & \small  Dice & \small  Jaccard   &\small  Dice& \small  Jaccard   &\small  Dice & \small Jaccard & \small  Dice\\ \hline\hline
				ResNet-151 &0.527  &0.690 & 0.304 & 0.466  & 0.144  &0.257 &0.149 &0.260 &0.125 &0.222 &0.245 & 0.379 \\
				ResNet-v2 & 0.539 & 0.706 & 0.310 & 0.473  & 0.159  & 0.274 &0.189 &0.318 &0.121 &0.216  &0.264 &0.397 \\
				DenseNet-169 & 0.538 & 0.699 & 0.324& 0.490  & 0.158  & 0.273 &0.213 &0.351 &0.134 &0.236  &0.273 &0.410 \\
				b0-EfficientNet & 0.554 & 0.713 & 0.324& 0.497  &0.157   & 0.272 &0.213 &0.351 &0.139 &0.242  &0.277 &0.415 \\
				\small {U-Eff (\textbf{TATL})}&\textcolor{blue}{0.565}  &\textcolor{red}{0.722}   &\textcolor{blue}{0.373}  & \textcolor{red}{0.549}  &0.157   &0.271  &0.268 &0.421 & 0.243&0.390 &0.321 &0.471  \\
				\small {L-Eff (\textbf{TATL})}  & 0.562  &0.719    &0.356  & 0.522  &\textcolor{blue}{0.168}   &\textcolor{red}{0.287}  &\textcolor{blue}{0.292} &\textcolor{red}{0.452 }&\textcolor{blue}{0.252} &\textcolor{red}{0.393}  &\textcolor{blue}{0.326} & \textcolor{red}{0.475}
				\\\hline
		\end{tabular}
		}
\end{table*}

We can see that all of the top results came from one of our techniques, with the U-shape architecture gaining first place in the Pigment Network and Globules features and the Link-shape architecture taking second. The baseline with the b0-EfficientNet backbone, on the other hand, appeared to outperform the other approaches. Considering skin characteristics with a large amount of training data, such as Pigment Network ($79.03\%$) or Milia-lie cysts ($33.55\%$), our model enhanced the Jaccard of the b0-EfficientNet from $55.4\%$ to $56.5\%$ and from $15.7\%$ to $16.8\%$, respectively. On Pigment Network, the Dice coefficient was improved by $0.9\%$, and on Milia-like cysts, it was improved by $1.6\%$. The lower the number of images, the greater the margin of improvement made by our model through the transfer learning step in TATL. For example, with the Streaks, our L-Eff(TATL) was $25.2\%$ and $39.3\%$ in Jaccard and Dice, correspondingly, which were $11.3\%$ and $15.1\%$ higher than the best baseline results with EfficientNet backbone.

{Overall, TATL with the Link-shape structure performed the best across all network backbones, followed by TATL with the U-shape with a minor margin. Furthermore, these configurations surpass all remaining baselines with a large margin, thereby proving the benefit of using TATL.} 

\subsubsection{Comparison of Network Parameters}
While achieving state-of-the-art performances, our method enjoys a significant reduction in the number of parameters. We provide the number of trainable parameters on different architectures in Table~\ref{tab:parameter}. Notably, compared to the winner of ISIC 2017 \citep{kawahara2018fully} and ISIC 2018 challenge \citep{koohbanani2018leveraging}, our method has {\bf 1.4 to 2.33 times} and {\bf 30 to 50 times} fewer parameters during training. Consequently, our TATL consumes less GPU memory and thus can be trained with higher image resolution or employed in mobile devices with low memory costs.

\begin{table}[!hbt]
\centering
\caption{{Number of parameters in each architecture in the ISIC challenge.}}
\vspace{0.1in}
\label{tab:parameter}
\scalebox{1.0}{
\begin{tabular}{|c|c|}
	\hline
	\textbf{Architecture} & \textbf{Number of Parameters}\\
	\hline\hline
	VGG16 & 138,357,544 \\
	ResNet-151 & 60,419,944	\\
	ResNet-v2  &55,873,736\\
	DenseNet-169 & 14,307,880\\
	EfficientNetb0 & 5,330,571\\
	ISIC2018-1st & 308,747,840\\
	ISIC2017-1st & 14,780,929 \\
\textbf{	Our (EfficientNet, U-shape)} & 10,115,501\\
\textbf{	Our (EfficientNet, Link-shape}) & 6,096,333\\
 	\hline
\end{tabular}}
\end{table}

\subsection{TATL Works Well with Various Network Architectures}
\label{sec:tatl-various-architectures}
In Section~\ref{sec:main-results}, we demonstrated that TATL achieved promising results using the b0-EfficientNet backbone. In this experiment, we explore the robustness of TATL to different network architectures beyond EfficientNets. Particularly, we compare ImageNet initialization against TATL on \emph{five} different backbone networks (VGG16, ResNet151, ResNet-v2, DenseNet-169, and EfficientNet-b0) and evaluate the performances on the Negative and Streaks attributes because they are the most challenging ones with the least training samples. In addition, we consider the following settings:

\begin{itemize}
    \item The first one, denoted as  \textit{TATL (FE)}, was to apply the TATL technique but froze the encoder part and only update weights of the decoder module while training for a specific skin attribute.
    \item The second configuration, denoted as \textit{TATL (NF)}, was similar to the former but allowed updating the parameters in the encoder.
    \item The last setting (\textit{ImageNet}) was not to apply the transfer learning process and train from scratch using weights pre-trained on the ImageNet dataset.
\end{itemize}

For each model, we use five different backbone architectures and two different convolution network shapes (U-shape and Link-shape). We also rerun three times for each configuration and measure the corresponding performance with the five-fold cross-validation technique to estimate average results. This configuration results in a total of {\bf $\mathbf{600}$ experiments} to be examined, which provides a comprehensive analysis of our TATL.

{Table~\ref{tab:result1718} reports the results of the experiment and shows that} applying TATL could help improve all backbone performance except the ResNet-v2 with the Negative attribute. However, the difference between the Dice values, in this case, was not noticeable (less than $1\%$). In contrast, the TATL could boost the Dice by nearly $13\%$ when using DenseNet-169 with U-shape to segment Streaks regions in the ISIC2018 dataset, and $11.4\%$ when using ResNet151 with Link-shape in a similar task. On the backbones such as DenseNet-169 + U-shape, ResNet151 + Link-shape, TATL consistently provided significant improvements. In summary, we find that our proposed TATL transfer learning could operate with various network architectures, thereby demonstrating its efficacy and generalizability.

\begin{table}
\centering
\caption{The average Dice coefficients of different backbones using our method and the transfer learning based on ImageNet using the datasets ISIC 2017 and 2018.  The bold and red-marked values are the highest and second-highest of each architecture. FE indicates freezing the encoder, NF indicates non-freezing and training both the encoder and decoders, L-Shape: Link-Shape.}
\vspace{0.1in}
\label{tab:result1718}
		\scalebox{1.0}{
\setlength\tabcolsep{3pt}
\begin{tabular}{|c|c|c|c|c|c|c|c|c|c|}
\hline
\multirow{3}{*}{\small \textbf{\small Architecture}} & \multirow{3}{*}{\textbf{\small Methods}}      & \multicolumn{4}{c|}{\small \textbf{ISIC 2017}}                                                                         & \multicolumn{4}{c|}{\small \textbf{ISIC 2018}}                                                                         \\ \cline{3-10}
&                                              & \multicolumn{2}{c|}{\small \textbf{Negative}}                 & \multicolumn{2}{c|}{\small \textbf{Streaks}}                  & \multicolumn{2}{c|}{\small \textbf{Negative}}                 & \multicolumn{2}{c|}{\small \textbf{Streaks}}                  \\ \cline{3-10} 
&                          & \textbf{\small U-shape}          & \textbf{\small L-Shape}        & \textbf{\small U-shape}           & \textbf{\small L-Shape}       & \textbf{\small U-shape}          & \textbf{\small L-Shape}        & \textbf{\small U-shape}           & \textbf{\small L-Shape}       \\ \hline \hline
                                            & \small TATL (FE) & \textbf{0.273}            & 0.178                      &  \textcolor{red}{ 0.281}  & 0.279                     &  \textcolor{red}{ 0.303} & 0..225                     & 0.266                      &  \textcolor{red}{ 0.272} \\  
                                            & \small TATL (NF)     &  \textcolor{red}{ 0.253} & 0.191                      & \textbf{0.282}             & 0.278                     & \textbf{0.309}            & 0.252                      & 0.262                      & \textbf{0.279}            \\ 
\multirow{-3}{*}{\small \textbf{\small VGG-16}}           & \small ImageNet              & 0.231                     & 0.232                      & 0.254                      & 0.225                     & 0.235                     & 0.244                      & 0.207                      & 0.254                     \\ \hline
                                            &\small  TATL (FE) & 0.231                     &  \textcolor{red}{ 0.279}  & 0.283                      &  \textcolor{red}{ 0.285} & \textbf{0.357}            & 0.33                       &  \textcolor{red}{ 0.317}  & 0.288                     \\  
                                            & \small TATL (NF)     & 0.248                     & \textbf{0.289}             & 0.281                      & \textbf{0.289}            &  \textcolor{red}{ 0.344} & 0.315                      & \textbf{0.324}             & 0.286                     \\  
\multirow{-3}{*}{\small \textbf{\small ResNet151}}        & \small ImageNet              & 0.239                     & 0.244                      & 0.175                      & 0.194                     & 0.275                     & 0.235                      & 0.201                      & 0.174                     \\ \hline
                                            & \small TATL (FE) & 0.294                     & 0.256                      &  \textcolor{red}{ 0.293}  & 0.248                     & 0.445                     & 0.406                      &  \textcolor{red}{ 0.324}  & 0.299                     \\  
                                            & \small TATL (NF)     & \textcolor{red}{0.30}                      & 0.279                      & \textbf{0.299}             & 0.252                     & \textcolor{red}{0.458}                     & 0.413                      & \textbf{0.329}             & 0.300                       \\  
\multirow{-3}{*}{\small \textbf{\small ResNet-v2}}        & \small ImageNet              & \textbf{0.313}            &  0.28   & 0.226                      & 0.198                     &   0.432 & \textbf{0.460}              & 0.235                      & 0.240                      \\ \hline
                                            & \small TATL (FE) & \textbf{0.339}            & 0.284                      & 0.299                      &  \textcolor{red}{ 0.303} & 0.292                     &  \textcolor{red}{ 0.367}  & 0.338                      & \textbf{0.368}            \\  
                                            & \small TATL (NF)     &  \textcolor{red}{ 0.307} & 0.288                      & 0.296 & \textbf{0.306}            & 0.353                     & \textbf{0.389}             & 0.342 &  \textcolor{red}{ 0.346} \\  
\multirow{-3}{*}{\small \textbf{\small DenseNet169}}      &\small  ImageNet              & 0.241                     & 0.227 & 0.205 & 0.194                     & 0.285                     & 0.377 & 0.210  & 0.216                     \\ \hline
                                            & \small TATL (FE) & 0.289                     & 0.295                      & \textbf{0.346}             & 0.321                     & 0.401                     & \textbf{0.445}                      & 0.384            & \textcolor{red}{0.395}                     \\  
                                            & \small TATL (NF)     &  \textcolor{red}{ 0.297} & \textbf{0.355}             &  \textcolor{red}{ 0.334}  & 0.332                     &   0.421 & \textcolor{red}{0.440}             &  0.359  & \textbf{0.410}                      \\  
\multirow{-3}{*}{\small \textbf{\small Eff. Net-b0}}  & \small ImageNet              & 0.286                     & 0.218                      & 0.259                      & 0.219                     & 0.355                     & 0.392                      & 0.199                      & 0.220                      \\ \hline
\end{tabular}}
\end{table}
\subsection{The Stability of TATL under The Influences of Data Size}
\begin{figure*}[!hbt]
\begin{center}
\includegraphics[width=0.9\textwidth]{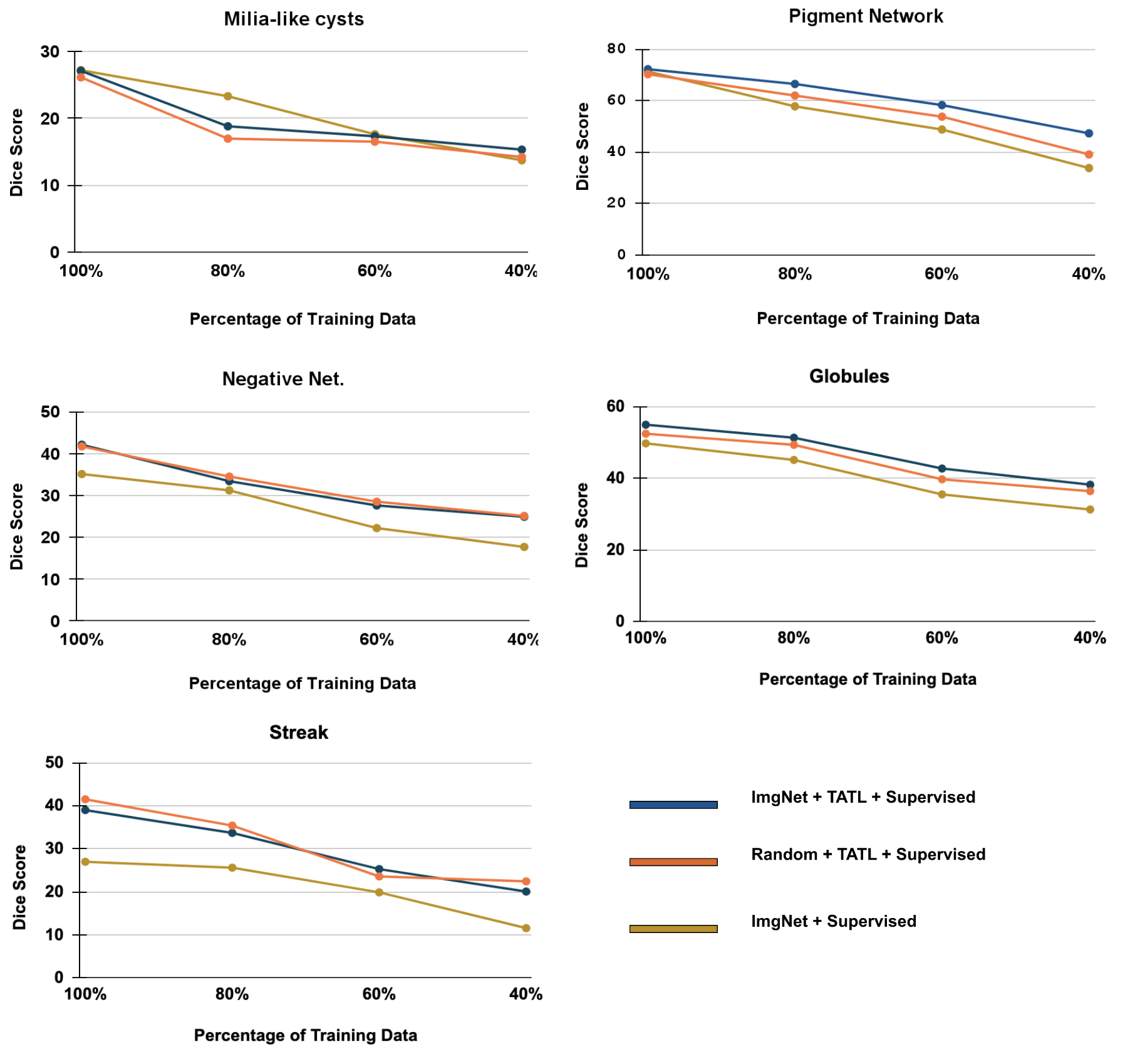}
\caption{The stability of TATL on five skin lesion attributes in ISIC 2018 under various sample sizes.}
\label{fig:stability}
\end{center}
\end{figure*}
Using pre-trained ImageNet models have been a common practice for many computer vision applications because of the diversity in the ImageNet dataset, making the pre-trained models stable and can detect local relationships. In contrast, TATL does not use additional data sources and only relies on the dataset at hand to create and learn the skin attribute regions. As a result, TATL's performance is subject to the amount of training data for the current task. Nevertheless, the results in Section~\ref{sec:main-results} showed that TATL had achieved promising results on standard benchmarks using all the labeled data provided. In this section, we explore the stability of TATL under the effect of {different} data size.

To test the stability of TATL, for each skin feature, its testing data are reserved for evaluation, and we subsample a portion of the remaining data for training. Particularly, we vary the amount of training data for each skin feature from 100\% (the original experiments in Section~\ref{sec:main-results}) to only 40\% and consider three competitors:
\begin{itemize}
    \item TATL using a pre-trained ImageNet model;
    \item TATL with a random initialization;
    \item Only a pre-trained ImageNet model.
\end{itemize}
We report the results in Figure~\ref{fig:stability}. In most cases, reducing the amount of training data results in worse performances for all methods. Moreover, both TATL versions achieved better performances than only using a pre-trained ImageNet model, except for the Millia feature at 80\% training data. Interestingly, the two TATL versions achieved similar performances on skin attributes with limited training data (Negative Net and Streak), while the gap was larger on the attributes with more training data (Pigment Network). However, overall, both TATL variants achieved better results than the standard strategy of using a pre-trained ImageNet. Furthermore, TATL with a pre-trained ImageNet model achieved the best-averaged results across attributes.

\subsection{Exploring TATL's Features Generality and Convergence Rate}
As discussed in Section~\ref{sec:inspiration}, our TATL is inspired by the dermatologists' behaviors, which motivates the learning of skin attribute regions. In this experiment, we examine the relationship between skin attribute regions' features learned by the Attribute-Agnostic Segmenter and the specific features learned for each particular attribute of Target-Segmenters. {We hypothesize that there are  additional benefits of performing supervised learning on attributes whose features are similar to those obtained from the Attribute-Agnostic learning step.} To validate this hypothesis, we consider two TATL variants: 
\begin{itemize}
    \item \emph{Downstream TATL}: The standard TATL framework {(Figure \ref{fig:training})}, where 
    the model for each skin attribute is initialized from the Attribute-Agnostic step;
    \item \emph{Pretext TATL}: A TATL variant which directly uses the model trained in the Attribute-Agnostic learning step to inference for each skin attribute, {\textit{without additional supervised learning in downstream task} (Figure \ref{fig:training} with only the left block)}.
\end{itemize}

\begin{figure*}[!hbt]
\begin{center}
\includegraphics[width=0.6\textwidth]{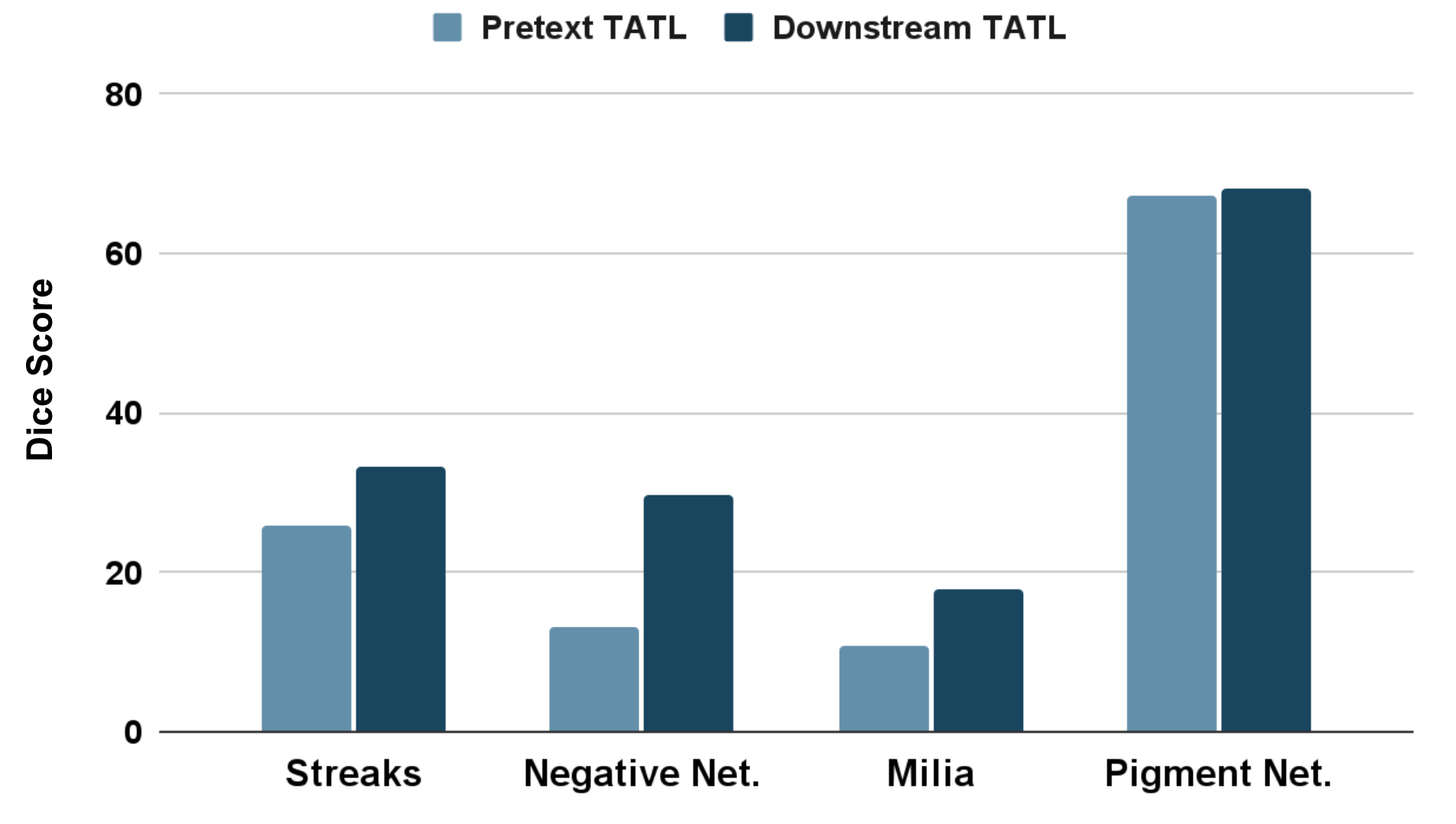}
\caption{The performance comparison of four skin lesion attributes in ISIC 2017 using (i) TATL trained directly on Pretext Task, and (ii) TATL after trained on the Downstream Task.}
\label{fig:compare-pretex-downstream}
\end{center}
\end{figure*}

\begin{figure*}[!hbt]
\begin{center}
\includegraphics[width=0.8\textwidth]{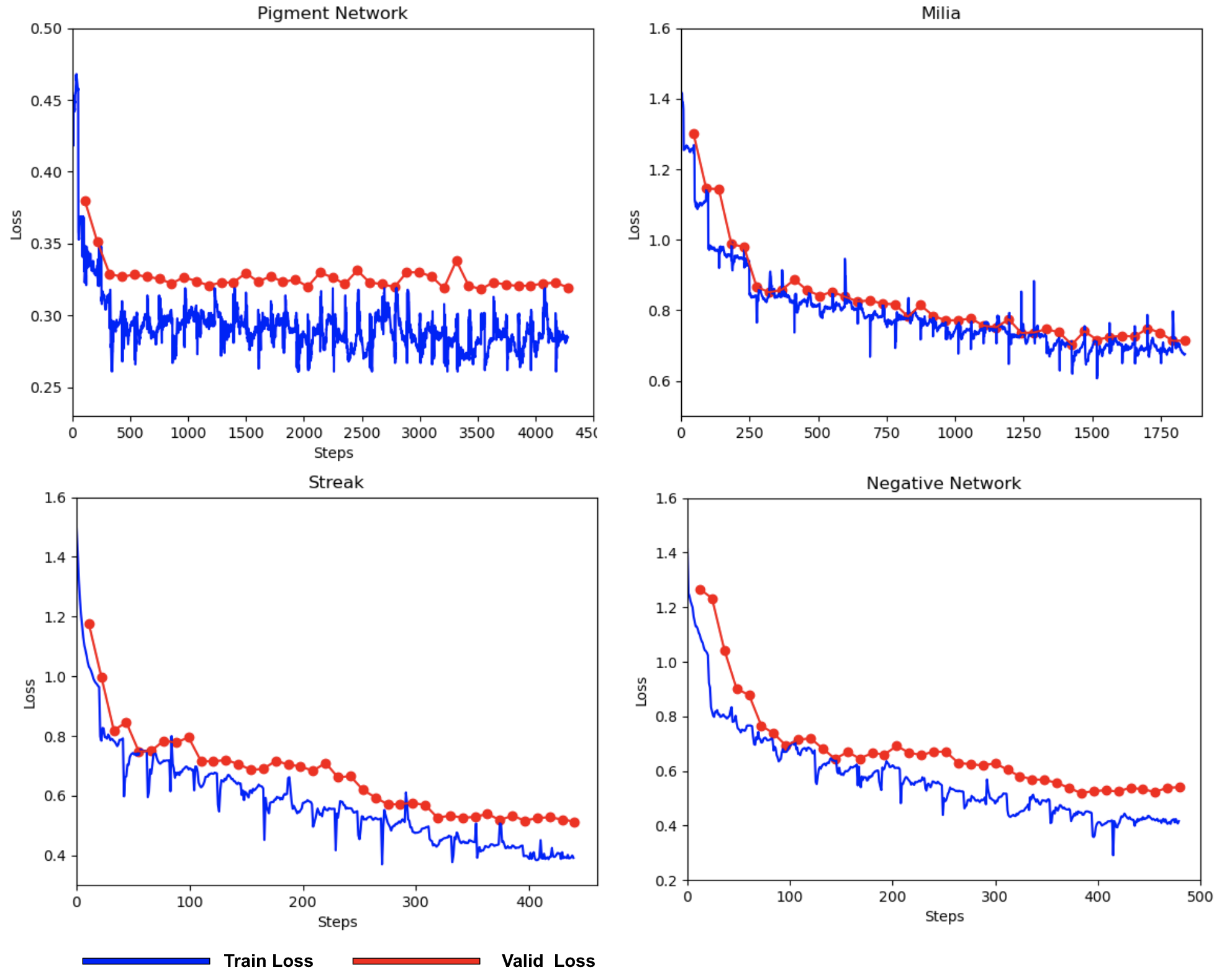}
\caption{The training and validation curves of four skin characteristics in ISIC 2017 with respect to the number of training steps.}
\label{fig:training_curve}
\end{center}
\end{figure*}

{We run experiments on the ISIC 2017 benchmark with the two aforementioned TATL variants and report the Dice score for each attribute after the learning procedures in Figure~\ref{fig:compare-pretex-downstream} as well as the loss curves in Figure~\ref{fig:training_curve} with respect to the number of training steps}. Note that the Pigment Network enjoys the most training samples amongst the four attributes, while the remaining three, Streaks, Negative Network, and Milia, are considered the minority classes.

Figure~\ref{fig:compare-pretex-downstream} shows that for the Pigment Network feature, the improvement of \textit{Downstream TATL} compared to \textit{Pretext TATL} is almost neglectable, which shows that the benefit of additional supervised training is minor for the attributes with ample training data. In contrast, the minor attributes' performance gaps are much more significant, suggesting TATL brings more benefits to the classes with limited training samples. We further verify this result in Figure~\ref{fig:training_curve} where the validation loss for the Pigment Network attribute plateaus very early on (at around 300 iterations) while it further decreases for other minor attributes. Overall, these experiments show that {the downstream fine-tuning step is particularly beneficial for the small classes. Therefore, one can infer that TATL, with all three stages, can significantly boost the performance on the classes with limited training data, which are more challenging to improve performance.}

\subsection{Comparison Between TATL and Other Training Paradigms}\label{sec:paradigms}
\begin{table}[!hbt]
\centering
\caption{Average performance over different skin attributes on ISIC-2018 with TATL compared to other training strategies. In each encoder-decoder type, Blue and Red colors are the best results in Jaccard and Dice coefficients, respectively.}
\vspace{0.1in}
\label{tab:TATL-against-initializers}
\scalebox{1.0}{
\begin{tabular}{|c|c|c|c|c|}
\hline
\multirow{2}{*}{\textbf{Methods}} & \multicolumn{2}{c|}{\textbf{U-shape}} & \multicolumn{2}{c|}{\textbf{Link-shape}} \\ \cline{2-5} 
                          & Dice         & Jaccard       & Dice          & Jaccard         \\ \hline \hline
\begin{tabular}[c]{@{}c@{}}TATL  \\ \end{tabular}
             & \textcolor{red}{0.471}       & \textcolor{blue}{0.321}         & \textcolor{red}{0.475 }        & \textcolor{blue}{0.326}           \\ \hline
\begin{tabular}[c]{@{}c@{}}Pretrained ImageNet\\ \end{tabular}           & 0.422        & 0.286         & 0.419         & 0.285           \\ \hline
\begin{tabular}[c]{@{}c@{}}Rotation-based SSL\\ \end{tabular} & 0.429       & 0.283         & 0.417         & 0.284           \\ 
\hline
\begin{tabular}[c]{@{}c@{}}Image Context-based SSL\\ \end{tabular}    & 0.446 & 0.298 & 0.426 &  0.301\\\hline
\begin{tabular}[c]{@{}c@{}}Data Augmentation\\ \end{tabular} & 0.435        & 0.294         & 0.403         & 0.276           \\ 
\hline
\begin{tabular}[c]{@{}c@{}} Attention-based Method\\ \end{tabular} & 0.441        & 0.296         & 0.416         & 0.285           \\ 
\hline
\end{tabular}}
\label{tab:comparing-initilizers}
\end{table}

In this experiment, we explore the benefits of TATL compared to other training paradigms, namely self-supervised learning (SSL), training with data augmentation, and attention-based methods. Note that our TATL is related to the SSL paradigm in terms of deriving better pre-trained models through solving auxiliary tasks.
For the SSL baselines, we implement  an additional pre-training phase to replace TATL's first and second stages.
Particularly, we consider the task of predicting the rotation angle applied on the input~\citep{DBLP:journals/corr/abs-1803-07728} or reconstructing the image after scrambling the pixels~\citep{DBLP:journals/corr/abs-1803-07728}. We also consider the strategy of supervised training with data augmentation to increase the total training samples, and using an attention-based architecture which can suppress irrelevant regions in an input image
while focusing salient features useful for a specific task~\citep{oktay2018attention}.

In summary, given the same U-shape architecture and pre-trained b0-EfficientNet as the network backbone,  we compare TATL against the standard Pre-trained ImageNet models and the following training strategies:
\begin{itemize}
    \item Self-supervised with the image rotation-based method \citep{DBLP:journals/corr/abs-1803-07728};
    \item Self-supervised with the image context restoration method \citep{chen2019self};
    \item Supervised training with data-augmentations, in which we use random rotation, flip, shift, brightness, or zoom;
    \item U-Eff network with attention gates as proposed in \citet{oktay2018attention}.
\end{itemize}

Table \ref{tab:comparing-initilizers} presents the experiment results with the metrics computed by averaging all skin attributes in the ISIC-2018 challenge.
In general, both the attention approach and the image reconstruction SSL can provide marginal improvements to the traditional Imagenet initialization on the U-Shape design. However, our TATL still outperforms such strategies on both evaluation metrics and architecture designs. This evidence confirms our finding that transferring knowledge from the Attribute-Agnostic Segmenter is beneficial for the skin-attribute segmentation task.

\subsection{Generalization Bound of TATL Compared to Other Strategies} \label{sec:compute_bound}
In Section~\ref{sec:paradigms}, we demonstrated that TATL could outperform standard training paradigms such as using a pre-trained model and SSL methods. This section explores how the theoretical insights developed in Section \ref{sec:theory} supports the empirical success of TATL. Recall that from Proposition~\ref{pro:proposition}, one can infer that given the same conditions of employed SGD algorithm and other hyper-parameters such as learning rate or the number of epochs, \textit{a model is expected to achieve small generalization errors on a testing set if its errors on the corresponding training data measured at the point of initialization (without any supervised learning) is small}. Given that TATL achieved lower testing errors in experiments (Table \ref{tab:result1718} and \ref{tab:TATL-against-initializers}), we now conduct a test to explicitly verify if this results correspond to a lowest TATL's generalization error bound values compared with other strategies. 

\begin{figure*}[!hbtp]
\begin{center}
\includegraphics[width=0.8\textwidth]{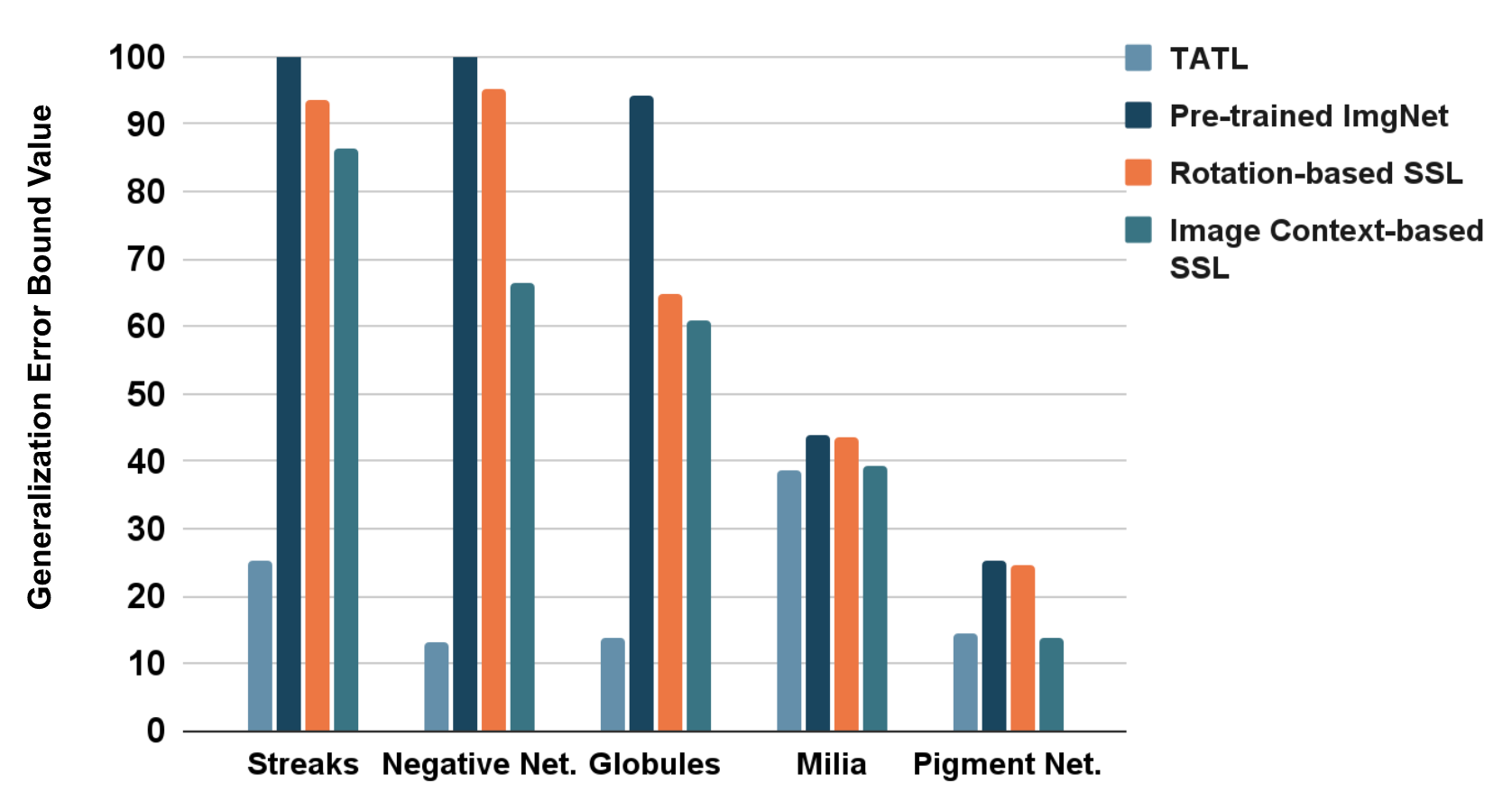}
\caption{Generalization error bound value (lower is better) in Proposition \ref{pro:proposition} measured on ISIC-2018 for different initialization methods. Values are scaled with a factor of $\mathbf{100}$ for better visualization.}
\label{fig:graph}
\end{center}
\end{figure*}

In particular, we estimate generalization error's bound in the right side of Eq. (\ref{eqn:prop1}) on the ISIC 2018 with four different cases: our TATL, pre-trained ImageNet, Rotation-based SSL \citep{DBLP:journals/corr/abs-1803-07728}, and Image Context Reconstruction-based SSL \citep{chen2019self}. We do not compare with the Attention-based method and Data Augmentation approach since both use the same Pre-trained ImageNet. For each method, we use the  U-Eff network and run a full pass over all training samples of each attribute to estimate the spectral norm of the Hessian matrix and the empirical risk $\hat{R}_{S}$. Here, the largest eigenvalue is approximated by the power iteration method \citep{solomon2015numerical}. We set $K=4,\, c = 0.01$ for all attributes and present the relative relations among categories in Figure~\ref{fig:graph}.

{It is noteworthy that our TATL acquired the lowest generalization error values for all skin attributes, especially with Streaks and Negative Net. These observations are compatible with our experiment, in which we outperformed other training strategies (as shown in Table \ref{tab:TATL-against-initializers}) and surpassed the Pre-trained ImageNet by a wide margin for the two characteristics Negative and Streak (Table \ref{tab:result1718}, at \textit{Eff. Net-b0} row and \textit{ISIC 2018} column). In conclusion, we argue that the TATL's efficacy could be demonstrated in both experimental and theoretical settings.}

\subsection{Visualization}
\begin{figure*}[!hbtp]
\begin{center}
\includegraphics[width=1.0\textwidth]{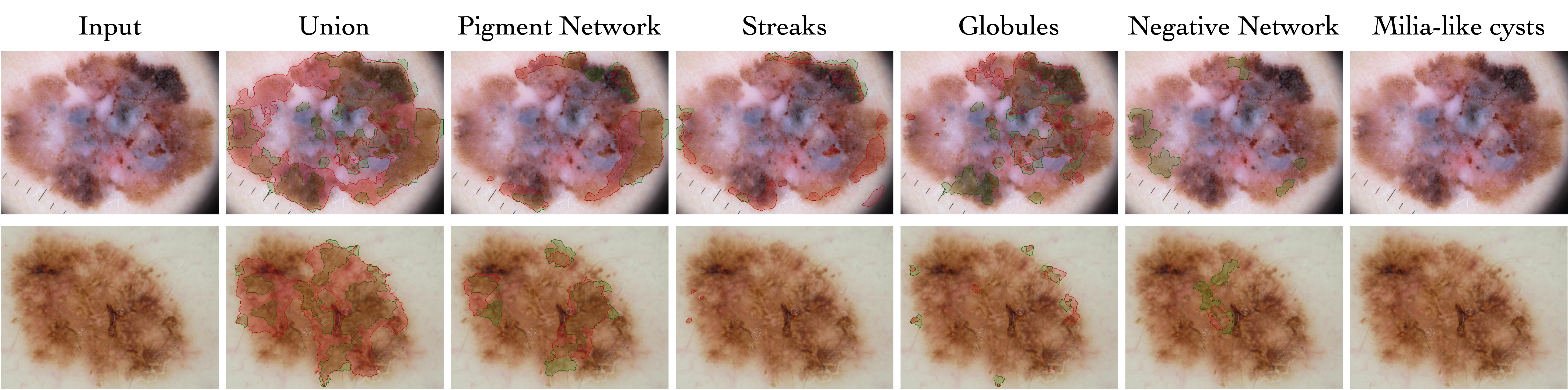}
\caption{An example of the TATL results in two typical skin images (zoom in for better view) where the green indicates the ground truth, and the red illustrates our prediction.}
\label{fig:visualize}
\end{center}
\end{figure*}

Figure \ref{fig:visualize} illustrates some sample results of our proposed TATL model. The ground-truth segmentation was highlighted in green, and our prediction was marked with red. Regarding attributes with many training samples such as Globules or Pigment Network, TATL has a better segmentation covering most ground truth areas. Furthermore, although Streaks and Negative Network's prediction missed some injured regions, the result still captured the primary matter location. 

\begin{figure*}[!hbtp]
\centering
\begin{tabular}{c}
\includegraphics[width=0.7\textwidth]{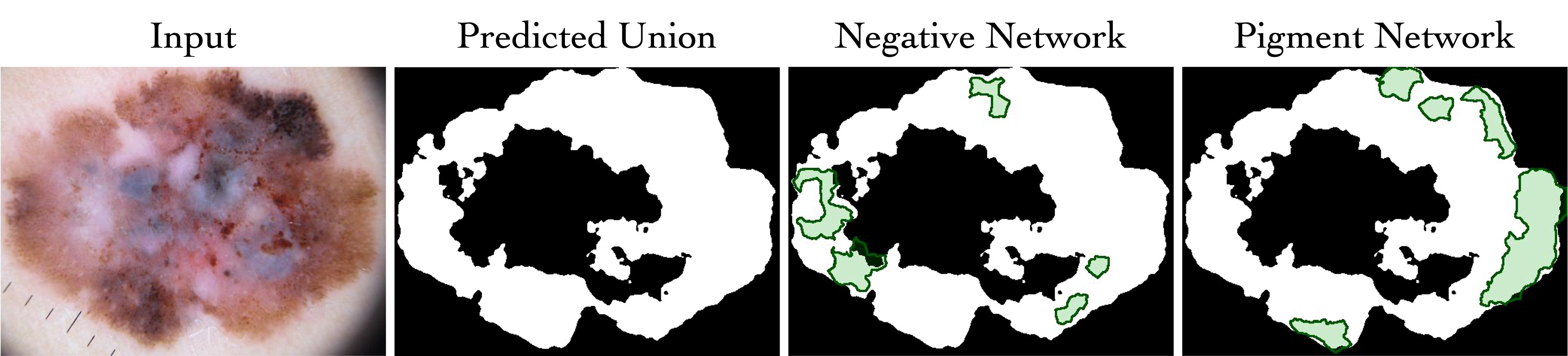} \\ [3pt]
\textbf{(a)} \\[1pt]
\end{tabular}
\begin{tabular}{cc}
\includegraphics[width=0.7\textwidth]{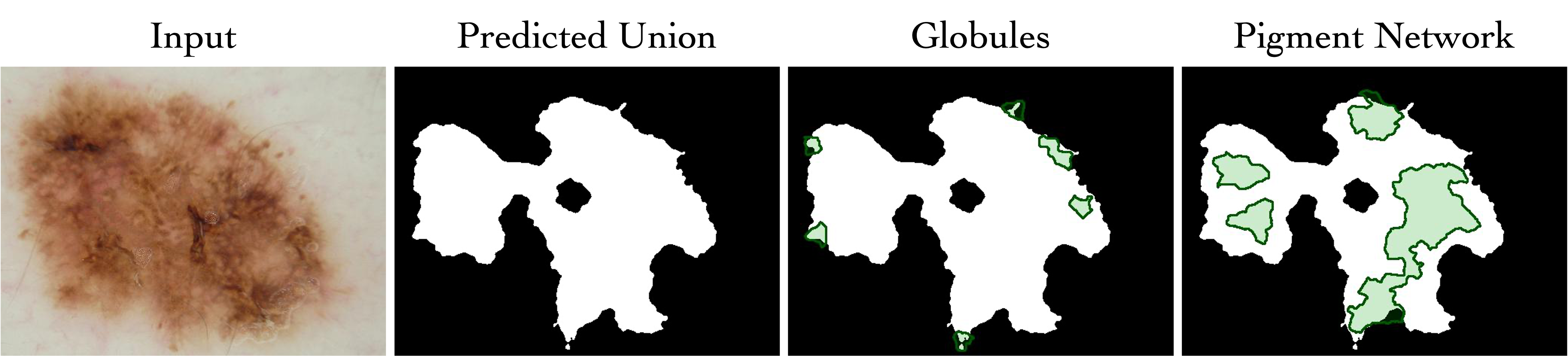} \\ [3pt]
\textbf{(b)}  \\[1pt]
\end{tabular}
\caption{{An illustration demonstrating the usefulness of our union segmentation, which can help dermatologists know which area they should focus on. Our predicted union can cover most of the ground truth of attribute regions which are depicted in the green areas.}}
\label{fig:overlap-union-gt}
\end{figure*}

{Our model also provides the benefit of extra information for end-users through the predicted union regions. For instance, we show, in Figure \ref{fig:overlap-union-gt}, the correlation between the union regions with typical skin attributes for the same example in Figure \ref{fig:visualize}. For each skin attribute, the ground truth is in green color, and we draw them over the binary maps to indicate the union positions' prediction. It can be seen that predicted union could cover both large regions as in Pigment Network and small disconnected regions as Negative Network (Figure \ref{fig:overlap-union-gt}a) and Globules (Figure \ref{fig:overlap-union-gt}b). 
This result can especially be useful under the scenarios where the dermatologists could not detect deformed or disconnected regions. In such cases, the union region provided by TATL can become helpful by highlighting the region of interest to assist the dermatologists.
As a result, TATL outputs not only can speed up the diagnosis process and but also help the dermatologists diagnose better, which is critical because they, not our model, will make the final diagnosis.}

\section{Discussion and Future Work}
Our work proposes a novel strategy to initialize the attribute segmenters' parameters using an attribute-agnostic segmenter trained on abnormal skin regions. We empirically demonstrate this benefit over the traditional strategy of using the ImageNet pretrained models. From the promising results, we outline several potential and interesting directions for future research. 

\paragraph{Generalization to Other Medical Image Analysis Tasks} We develop TATL to address the skin-attribute detection problem specifically. It would be interesting to test the TATL's generalization capabilities to other medical image analysis tasks, where using pretrained Imagenet models is likely to be suboptimal. For example, similar tasks such as brain lesion segmentation \citep{hu2018deep, duy2018accurate,mallick2019brain, nguyen20173d}, abnormal chest detection \citep{hashir2020quantifying, ibrahim2021pneumonia,nguyen2020attention}, or diabetic retinopathy lesion segmentation share similar characteristics to our problem setting: the data are often imbalance and classes share semantic features that can be leveraged to improve the overall performance. Therefore, it is of interest to explore the applications of TATL in such tasks and make possible adjustments.

\paragraph{Real-world Applications Using TATL} Our ultimate goal is to develop a model that not only makes predictions but also provides helpful information and assists {dermatologists} in making the final decisions. Our TATL framework realizes this goal by providing a mask of skin attribute regions regardless of their attributes, compensating for inaccurate predictions of later stages, especially on minor classes. A promising future direction for TATL is integrating it in an online learning setting with human-in-the-loop \citep{nunnari2021software,nunnari2021anomaly}. Particularly, a model is trained to detect some diseases and then deployed to a real-world environment with a stream of data and feedback from dermatologists and patients. In such scenarios, the model can continuously improve its performance by accumulating the attribute-agnostic information via the dermatologists' feedback and then transferring it to the target segmenters, allowing for a fast adaptation to newer patients and more accurate predictions over time.

\paragraph{A Holistic Medical Image Analysis Method Beyond TATL} Intuitively, TATL works by achieving a tighter generalization error bound compared to other initialization strategies. However, the theoretical result in Proposition~\ref{pro:proposition} only bounds using the initialization parameters. In practice, additional aspects can affect the model's generalization, such as (i) the number of source tasks (training classes in our case); (ii) which properties among those tasks that can be safely transferred; and (iii) beyond an initialization, which mechanisms allow for a successful knowledge transfer. Such properties are not yet rigorously studied, and exploring them can potentially provide a holistic method for medical image analysis: a method not only starts with a quality initialization but also exploits the complex relationship of medical images to improve its performance over time. Such a method can provide accurate detection and assist {dermatologists} in diagnosing rare diseases more precisely, which results in effective treatments at a lower cost.
\section{Conclusion}\label{concl}
We have investigated the limitations of the common fine-tuning strategy in state-of-the-art skin attributes detection methods. We show that such strategies are not optimal when the current task is largely different from ImageNet and contains limited training data. This limitation motivated us to develop TATL, a novel transfer learning method that exploits all attribute data to train the agnostic segmenter. By transferring the agnostic segmenter's knowledge to each attribute classifier, TATL alleviates issues of training data scarcity, especially for small classes, and allows knowledge sharing among attribute models. Through extensive experiments on the ISIC 2017 and ISIC 2018 benchmarks, we demonstrate the efficacy of TATL over existing state-of-the-art methods. Moreover, TATL is proven to work effectively with various backbone networks while enjoying minimal model and computational complexity. Finally, we present theoretical insights that demonstrate that TATL works in practice by bridging the domain gap via the task-agnostic segmenter, thus leading to competitive performance.
\section{Acknowledgement}
This research has been supported by the Ki-Para-Mi project (BMBF, 01IS1903-8B), the pAItient project (BMG, 2520DAT0P2), and the Endowed Chair of Applied Artificial Intelligence, Oldenburg University. Binh T. Nguyen is funded by Vietnam National University Ho Chi Minh City (VNU-HCM) under grant number NCM2019-18-01. We would like to thank Dr. Fabrizio Nunnari (German Research Centre for Artificial Intelligence, Germany) and Dr. Paul Swoboda (Max Planck Institute for Informatics, Germany) for their valuable discussions.

\bibliographystyle{model2-names.bst}\biboptions{authoryear}
\bibliography{mybibfile}

\newpage
\appendix
\section{Additional Experiment Results}
\begin{figure*}[!hbtp]
\begin{center}
\includegraphics[width=0.85\textwidth]{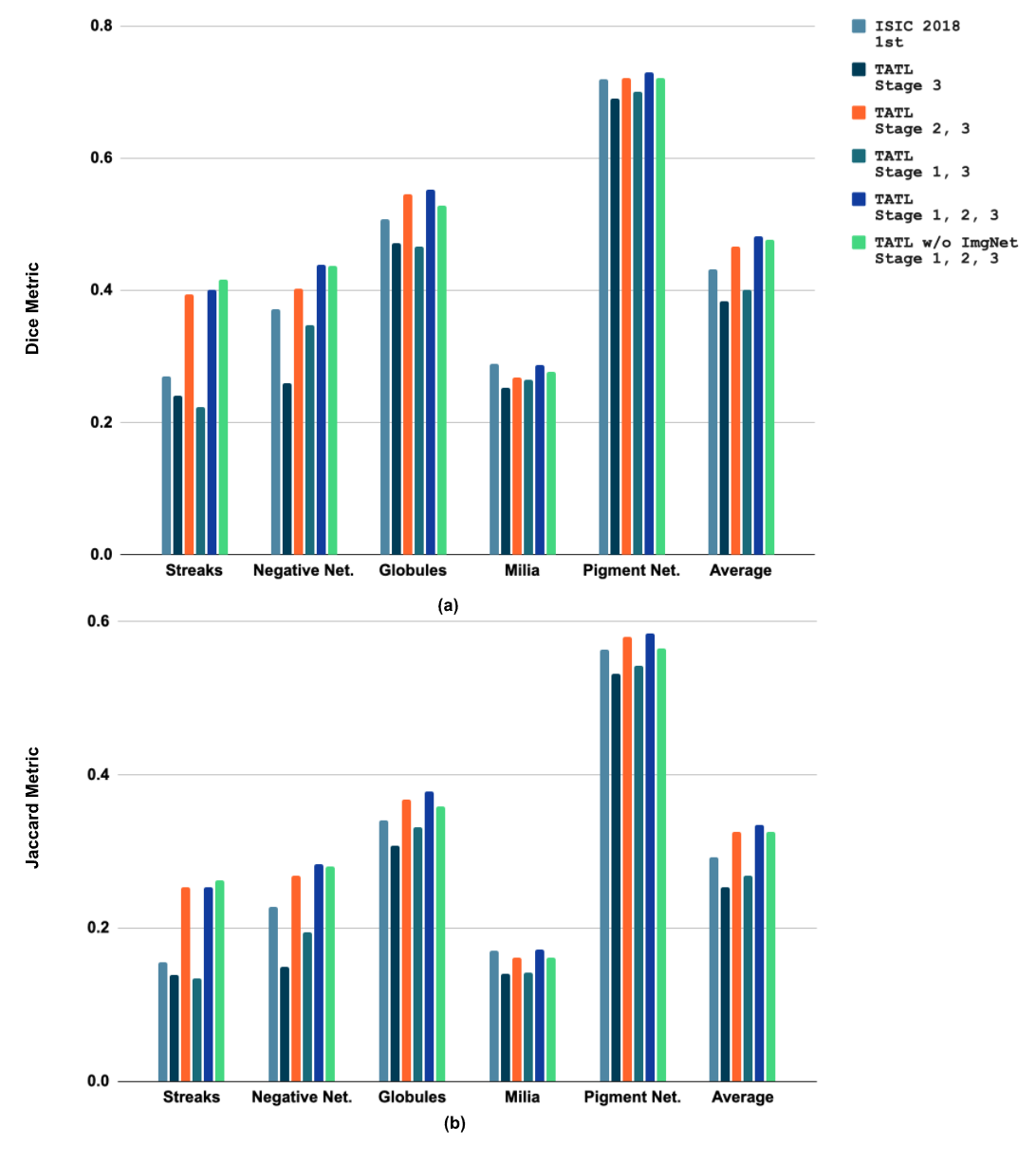}
\caption{ISIC-2018 Challenge Performance Visualization with respect to results in Table \ref{tab:our_result2018}}
\label{fig:isic-2018-chart}
\end{center}
\end{figure*}

\begin{figure*}[!hbtp]
\begin{center}
\includegraphics[width=0.8\textwidth]{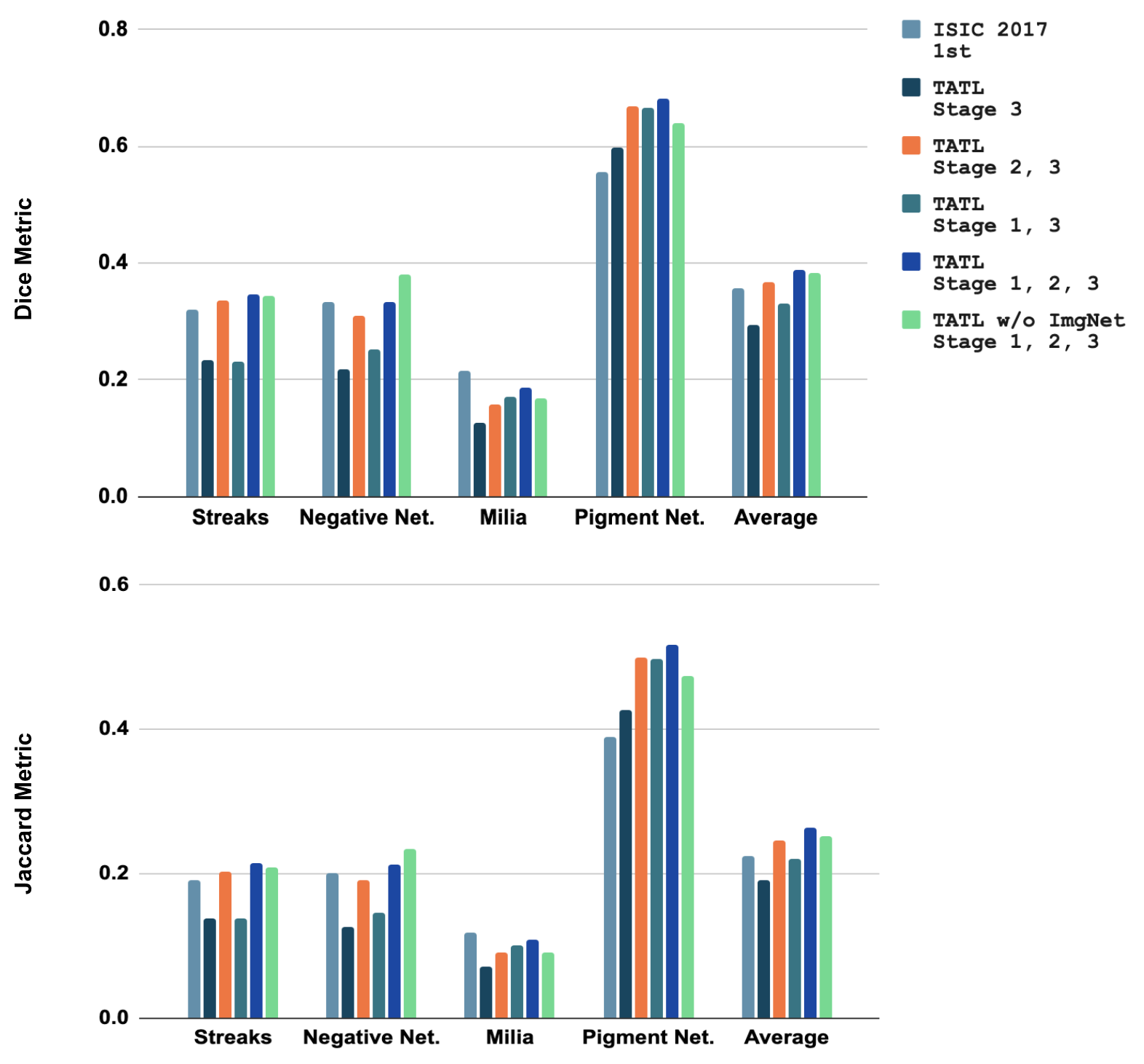}
\caption{ISIC-2017 Challenge Performance Visualization with respect to results in Table \ref{tab:our_result2017}}
\label{fig:isic-2017-chart}
\end{center}
\end{figure*}

This Appendix provides additional results of our experiments in Section~\ref{sec:main-results}.
Particularly, Figures~\ref{fig:isic-2018-chart} and~\ref{fig:isic-2017-chart} provide a visual comparison amongst different methods on the ISIC 2018 and ISIC 2017 challenges, respectively. Lastly, Tables~\ref{tab:jaccard_result2018-full-std}, \ref{tab:dice_result2018-full-std}, and \ref{tab:jaccard_result2017-full-std}, \ref{tab:dice_result2017-full-std} provide the standard deviation of the Jaccard and Dice on the ISIC 2018 and 2017 challenges respectively.

\begin{table*}[!hbt]
\centering
\caption{Standard deviation of the Jaccard metric on the ISIC2018 challenge.  The best results are in bold font. Stage 1: segmenting the lesion region, Stage 2: training the Attribute-Agnostic Segmenter, Stage 3: training the Target-Segmenters.}
\vspace{0.1in}
\label{tab:jaccard_result2018-full-std}
    \setlength\tabcolsep{3pt}
    \centerline{
	\begin{tabular}{|c|c|c|c|c|c|c|}
		\hline
		\textbf{Method}    & {  ISIC2018 1st}   & { \begin{tabular}[c]{@{}c@{}}TATL \\\textbf{Stage 3} \end{tabular}}& {\begin{tabular}[c]{@{}c@{}} TATL \\ \textbf{Stage 2, 3}\end{tabular}} & {\begin{tabular}[c]{@{}c@{}}  TATL \\ \textbf{Stage 1, 3}\end{tabular}} & {\begin{tabular}[c]{@{}c@{}}  TATL \\ \textbf{Stage 1, 2, 3}\end{tabular}} &
		{\begin{tabular}[c]{@{}c@{}}  TATL w/o ImgNet\\ \textbf{Stage 1, 2, 3} \end{tabular}} 
 \\ \hline\hline
		\textbf{Pigment Net.} & 0.563  &0.532 $\pm 0.016$    &0.580 $\pm 0.017$   &0.542 $\pm 0.016$ &\textbf{0.584 $\mathbf{\pm 0.016}$} &0.565 $\pm 0.017$ \\
		\textbf{Globules} & 0.341  &0.308  $\pm 0.023$  &0.368   $\pm 0.025$  &0.332 $\pm 0.024$ &\textbf{0.379 $\mathbf{\pm 0.023}$}   &0.359 $\pm 0.023$\\
		\textbf{Milia-like cysts} & \textcolor{black}{0.171}  &0.141 $\pm 0.016$  &0.161 $\pm 0.015$ &0.142 $\pm 0.014$ &\textbf{0.172 $\mathbf{\pm 0.015}$} & 0.161 $\pm 0.015$ \\
		\textbf{Negative Net.}& 0.228   &0.149 $\pm 0.013$ &\textcolor{black}{0.269 $\pm 0.014$}     &0.194 $\pm 0.015$  &\textbf{0.283 $\mathbf{\pm 0.013}$}  & 0.280 $\pm 0.014$ \\
		\textbf{Streaks}  & 0.156   &0.139 $\pm 0.044$   & 0.254  $\pm 0.042$   &0.135 $\pm 0.043$ &0.254 $\pm 0.045$ & \textbf{0.263 $\mathbf{\pm 0.047}$ }
		 \\ \hline
		\textbf{Average}  & 0.292   &0.254 $\pm 0.022$ &0.326  $\pm 0.023$ &0.269 $\pm 0.022$ &\textbf{0.334 $\mathbf{\pm 0.022}$}  & 0.326 $\pm 0.023$
		\\\hline
    \end{tabular}
    }
\end{table*}

\begin{table*}[!hbt]
\centering
\caption{Standard deviation of the Dice metric on the ISIC2018 challenge.  The best results are in bold font. Stage 1: segmenting the lesion region, Stage 2: training the Attribute-Agnostic Segmenter, Stage 3: training the Target-Segmenters.}
\vspace{0.1in}
\label{tab:dice_result2018-full-std}
    \setlength\tabcolsep{3pt}
    \centerline{
	\begin{tabular}{|c|c|c|c|c|c|c|}
		\hline
		\textbf{Method}    & {  ISIC2018 1st}   & { \begin{tabular}[c]{@{}c@{}}TATL \\\textbf{Stage 3} \end{tabular}}& {\begin{tabular}[c]{@{}c@{}} TATL \\ \textbf{Stage 2, 3}\end{tabular}} & {\begin{tabular}[c]{@{}c@{}}  TATL \\ \textbf{Stage 1, 3}\end{tabular}} & {\begin{tabular}[c]{@{}c@{}}  TATL \\ \textbf{Stage 1, 2, 3}\end{tabular}} &
		{\begin{tabular}[c]{@{}c@{}}  TATL w/o ImgNet\\ \textbf{Stage 1, 2, 3} \end{tabular}} 
 \\ \hline\hline
		\textbf{Pigment Net.}   & 0.720   &0.691 $\pm 0.013$   &0.721 $\pm 0.013$   &0.701  $\pm 0.013 $ &\textbf{0.730  $\mathbf {\pm 0.014}$}  & 0.721 $\pm 0.014$\\
		\textbf{Globules}  & 0.508 &0.471 $\pm 0.024$  &\textcolor{black}{0.546 $\pm 0.025$}  &0.467 $\pm 0.027$  &\textbf{0.552 $\mathbf{\pm 0.024}$}  & 0.528 $\pm 0.025$\\
		\textbf{Milia-like cysts}  & \textbf{0.289} &0.252 $\pm 0.024$  &0.268 $\pm 0.025$ &0.264 $\pm 0.025$  &\textcolor{black}{0.288 $\pm 0.025$}  & 0.277 $\pm 0.027$ \\
		\textbf{Negative Net.}  & 0.371   &0.260 $\pm 0.015$    &\textcolor{black}{0.403 $\pm 0.014$}   &0.348 $\pm 0.015$  & \textbf{0.438 $\mathbf{\pm 0.015}$} &  0.437 $\pm 0.016$\\
		\textbf{Streaks}   & 0.270   &0.241 $\pm 0.062$   &\textcolor{black}{0.394 $\pm 0.058$}  &0.224 $\pm 0.062$ &0.401 $\pm 0.061$ &  \textbf{0.416 $\mathbf{\pm 0.064}$}
		 \\ \hline
		\textbf{Average}   & 0.432   &0.383 $\pm 0.028$   &0.466 $\pm 0.027$ &0.401 $\pm 0.028$ & \textbf{0.482 $\mathbf{\pm 0.028}$} & 0.476 $\pm 0.029$
		\\\hline
    \end{tabular}
    }
\end{table*}

\begin{table*}[!hbt]
\centering
\caption{Standard deviation of the Jaccard metric on the ISIC2017 challenge. The best results are in bold font. Stage 1: segmenting the lesion region, Stage 2: training the Attribute-Agnostic Segmenter, Stage 3: training the Target-Segmenters.}
\vspace{0.1in}
\label{tab:jaccard_result2017-full-std}
    \setlength\tabcolsep{3pt}
    \centerline{
	\begin{tabular}{|c|c|c|c|c|c|c|}
		\hline
		\textbf{Method}    & {  ISIC2017 1st}   & { \begin{tabular}[c]{@{}c@{}}TATL \\\textbf{Stage 3} \end{tabular}}& {\begin{tabular}[c]{@{}c@{}} TATL \\ \textbf{Stage 2, 3}\end{tabular}} & {\begin{tabular}[c]{@{}c@{}}  TATL \\ \textbf{Stage 1, 3}\end{tabular}} & {\begin{tabular}[c]{@{}c@{}}  TATL \\ \textbf{Stage 1, 2, 3}\end{tabular}} &
		{\begin{tabular}[c]{@{}c@{}}  TATL w/o ImgNet\\ \textbf{Stage 1, 2, 3} \end{tabular}} 
 \\ \hline\hline
		\textbf{Pigment Net.} & 0.389   &0.426 $\pm 0.035$   &0.499 $\pm 0.033$ &0.497 $\pm 0.037$   & \textbf{0.516 $\mathbf{\pm 0.036}$}   & 0.473  $\pm 0.035$\\
		\textbf{Milia-like cysts} & \textbf{0.119}  &0.072 $\pm 0.025$ &0.091 $\pm 0.028$ &0.101 $\pm 0.025$  &0.108  $\pm 0.024$ & 0.092 $\pm 0.029$ \\
		\textbf{Negative Net.}& 0.201  &0.126 $\pm 0.044$   &0.191 $\pm 0.037$  &0.147 $\pm 0.038$  &0.213 $\pm 0.033$  & \textbf{0.234 $\mathbf{\pm 0.035}$} \\
		\textbf{Streaks}  & 0.192    &0.139 $\pm 0.033$    &0.203 $\pm 0.045$  &0.139 $\pm 0.037$   &\textbf{0.215 $\mathbf{\pm 0.039}$} & 0.209 $\pm 0.050$ \\ \hline
		\textbf{Average}  & 0.225   &0.191 $\pm 0.034$ &0.246 $\pm 0.036$ &0.221 $\pm 0.034$ &\textbf{0.263 $\mathbf{\pm 0.033}$}  & 0.252 $\pm 0.037$
		\\\hline
    \end{tabular}}

\end{table*}

\begin{table*}[!hbt]
\centering
\caption{Standard deviation of the Dice metric on the ISIC2017 challenge.  The best results are in bold font. Stage 1: segmenting the lesion region, Stage 2: training the Attribute-Agnostic Segmenter, Stage 3: training the Target-Segmenters.}
\vspace{0.1in}
\label{tab:dice_result2017-full-std}
    \setlength\tabcolsep{3pt}
    \centerline{
	\begin{tabular}{|c|c|c|c|c|c|c|}
		\hline
		\textbf{Method}    & {  ISIC2017 1st}   & { \begin{tabular}[c]{@{}c@{}}TATL \\\textbf{Stage 3} \end{tabular}}& {\begin{tabular}[c]{@{}c@{}} TATL \\ \textbf{Stage 2, 3}\end{tabular}} & {\begin{tabular}[c]{@{}c@{}}  TATL \\ \textbf{Stage 1, 3}\end{tabular}} & {\begin{tabular}[c]{@{}c@{}}  TATL \\ \textbf{Stage 1, 2, 3}\end{tabular}} &
		{\begin{tabular}[c]{@{}c@{}}  TATL w/o ImgNet\\ \textbf{Stage 1, 2, 3} \end{tabular}} 
 \\ \hline\hline
		\textbf{Pigment Net.} & 0.556  &0.597 $\pm 0.028$ &0.667 $\pm 0.033$  & 0.665 $\pm 0.031$ &\textbf{0.681 $\mathbf{\pm 0.034}$} & 0.639 $\pm 0.030$ \\
		\textbf{Milia-like cysts}  & \textbf{0.215}  &0.127 $\pm 0.032$  &0.157 $\pm 0.037$ &0.172 $\pm 0.035$ &0.188 $\pm 0.035$ & 0.168 $\pm 0.034$\\
		\textbf{Negative Net.}  & 0.333    &0.218 $\pm 0.037$  & 0.310 $\pm 0.038$ & 0.251 $\pm 0.046$  &0.334 $\pm 0.039$ &  \textbf{0.380 $\mathbf{\pm 0.036}$}\\
		\textbf{Streaks}   & 0.321   &0.233 $\pm 0.047$   &0.336 $\pm 0.062$   &0.232 $\pm 0.049$  &\textbf{0.346 $\mathbf{\pm 0.054}$} & 0.345 $\pm 0.066$ \\ \hline
		\textbf{Average}   &  0.356   & 0.294 $\pm 0.036$ & 0.367 $\pm 0.043$ &0.330 $\pm 0.040$ & \textbf{0.387 $\mathbf{\pm 0.041}$} & 0.383 $\pm 0.042$
		\\\hline
    \end{tabular}
}
\end{table*}

\end{document}